\documentclass[10pt,twocolumn,letterpaper]{article}

\usepackage{cvpr}              % To produce the CAMERA-READY version
\definecolor{cvprblue}{rgb}{0.21,0.49,0.74}
\usepackage[pagebackref,breaklinks,colorlinks,allcolors=cvprblue]{hyperref}
\usepackage{booktabs}
\usepackage{multirow}
\usepackage{tabularx}
\usepackage{float}
\usepackage{listings}
\usepackage{xcolor}
\usepackage{multirow}
\usepackage{booktabs}
\usepackage{textcomp}
\usepackage{verbatim}
\usepackage{tocloft}
\usepackage{appendix}
\usepackage{minitoc}

\definecolor{codegreen}{rgb}{0,0.6,0}
\definecolor{codegray}{rgb}{0.5,0.5,0.5}
\definecolor{codepurple}{rgb}{0.58,0,0.82}
\definecolor{backcolour}{rgb}{0.95,0.95,0.92}

\lstdefinestyle{mystyle}{
    backgroundcolor=\color{backcolour},   
    commentstyle=\color{codegreen},
    keywordstyle=\color{magenta},
    numberstyle=\tiny\color{codegray},
    stringstyle=\color{codepurple},
    basicstyle=\ttfamily\footnotesize,
    breakatwhitespace=false,         
    breaklines=true,                 
    captionpos=b,                    
    keepspaces=true,                 
    numbers=left,                    
    numbersep=5pt,                  
    showspaces=false,                
    showstringspaces=false,
    showtabs=false,                  
    tabsize=2
}

\def\paperID{18653} % *** Enter the Paper ID here
\def\confName{CVPR}
\def\confYear{2026}

\newcommand{\myparagraph}[1]{\vspace{0.3em}\noindent\textbf{#1}}

\newcolumntype{b}{X}
\newcolumntype{s}{>{\hsize=.5\hsize}X}
\newcolumntype{t}{>{\hsize=.2\hsize}X}

\def\ourmethod{SEAM }
\def\ourmethodnospace{SEAM}
\title{Rethinking Intermediate Representation for VLM-based Robot Manipulation}

\author{Weiliang Tang$^{1*}$, 
Jialin Gao$^{1*}$,
Jia-Hui Pan$^{1}$,
Gang Wang$^{1}$, 
Li Erran Li$^{2}$, \\
Yunhui Liu$^{1}$, 
Mingyu Ding$^{3}$,
Pheng-Ann Heng$^{1\dagger}$,
Chi-Wing Fu$^{1\dagger}$
\vspace{0.2cm} \\
  $^1$CUHK, \quad
  $^2$Amazon, \quad
  $^3$UNC 
  \vspace{-0.5cm}
}

\begin{document}
\twocolumn[{%
\maketitle

\begin{figure}[H]
\hsize=\textwidth
    \centering
    \includegraphics[width=0.98\textwidth]{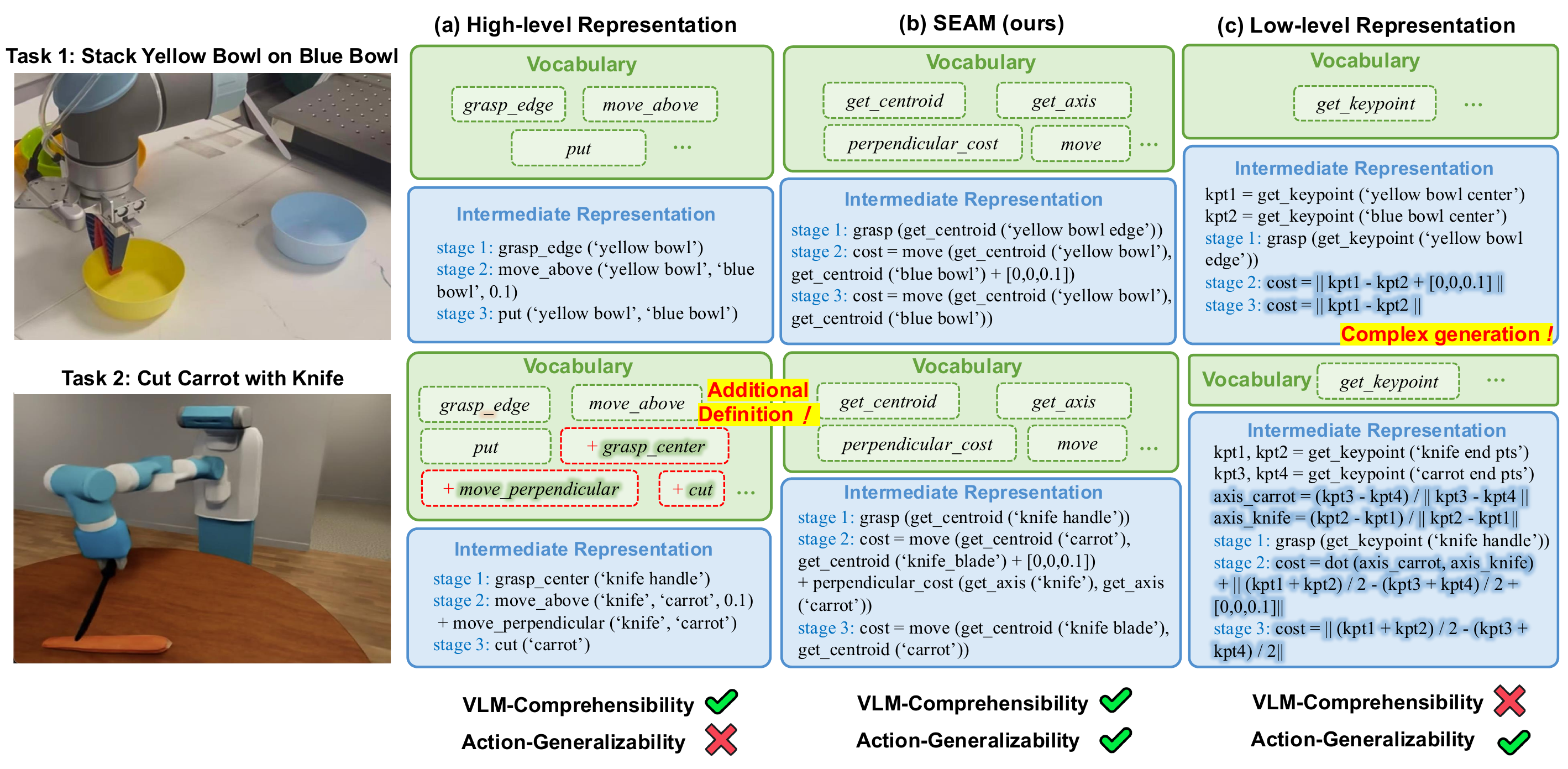}
    \vspace*{-2mm}
    \caption{Comparing (a) high-level, (c) low-level, and (b) %our \ourmethod (Composable Semantic) 
    our \ourmethod ({\underline {Se}}mantic {\underline A}sse{\underline m}bly) representation for supporting robot manipulation.
    High-level representation requires manually adding new vocabulary words to customize the model for new tasks, despite its VLM-comprehensibility, whereas low-level representation requires generating complex constraints in task handling, despite its generalizability in robot actions.
    Our new \ourmethod design meets the goals of both VLM-comprehensibility and action-generalizability.
    }
    \label{fig:teaser}
\end{figure}
}]

\begin{abstract}
Vision-Language Model (VLM) is an important component to enable robust robot manipulation. 
Yet, using it to translate human instructions into an action-resolvable intermediate representation often needs a tradeoff between VLM-comprehensibility and generalizability.
Inspired by context-free grammar, we design the \textbf{Semantic Assembly} representation named \ourmethodnospace, by decomposing the intermediate representation into vocabulary and grammar.
Doing so leads us to a concise vocabulary of semantically-rich operations and a VLM-friendly grammar for handling diverse unseen tasks. 
In addition, we design a new open-vocabulary segmentation paradigm with a retrieval-augmented few-shot learning strategy to 
%precisely 
localize fine-grained object parts for manipulation, 
%(e.g., cup handle, teapot opening) 
effectively with the \textbf{shortest} inference time over all state-of-the-art parallel works.
Also, we formulate new metrics for action-generalizability and VLM-comprehensibility, demonstrating the compelling performance of \ourmethod over mainstream representations on both aspects.
Extensive real-world experiments further manifest its SOTA performance under varying settings and tasks.
\end{abstract} 

\addtocontents{toc}{\protect\setcounter{tocdepth}{-1}}

\vspace*{-3mm}
\section{Introduction}
\label{sec:intro}

\vspace{1em}
\noindent \textit{``The limits of my language mean the limits of my world.''}\\
\hspace*{13em}--- \textit{Ludwig Wittgenstein}
\newline

Leveraging the immense knowledge in Vision-Language Models (VLMs) offers a new path toward general robot manipulation.
Typically, VLMs are employed to translate human instructions into an intermediate representation, specified,~\eg, by predefined action words and spatial constraints, such that we can solve to derive the robot actions.
This approach reduces the burden of preparing extensive annotations for training vision-language action (VLA) models to align robot actions with pretrained knowledge.

Earlier VLM-based methods~\cite{liang2023code, ahn2022can, huang2023instruct2act, liu2025hycodepolicy, wang2023demo2code,liu2024relep} design {\bf high-level\/} predefined skill words (see the example vocabularies in Fig.~\ref{fig:teaser}(a)) for constructing the intermediate representation.
Though the skill words are easy for the VLMs to comprehend and manipulate, the construction is overly rigid.
So, this approach often scales poorly to handle diverse and unseen tasks,~\eg, from \textit{``Stack Yellow Bowl on Blue Bowl''} to \textit{``Cut Carrot with Knife''}, and requires a tedious process of manually designing new skills,~\ie, \textit{``grasp\_center''}, \textit{``Cut''}, and \textit{``move\_perpendicular''}.

Recent studies have turned to a \textbf{low-level} representation approach by considering fundamental primitives such as keypoint and axes~\cite{huang2024rekep, pan2025omnimanip, huang2023voxposer, pan2025omnimanip,huang2024copa, tang2025geomanip}); see,~\eg, Fig.~\ref{fig:teaser}(c).
This approach offers greater flexibility and potential for cross-task generalization.
However, the resulting intermediate representations are often overly complex, thereby difficult for VLMS to comprehend and generate,~\eg, the task ``cutting the carrot using a grasped knife'' requires ReKep~\cite{huang2024rekep} to generate highly specific code to explicitly calculate the constraints and costs, which is not robust.
Driven by the above observations, we ask this question: \textit{How can we create an intermediate representation that can be (i) generalizable for diverse real-world tasks \textbf{(Action-generalizability)} and also (ii) sufficiently understandable for VLMs to reason \textbf{(VLM-comprehensibility)}?}

Drawing insights from context-free grammar, 
we design {\bf \ourmethodnospace\/}, a new 
\textbf{{\underline{Se}}mantic {\underline A}sse{\underline m}bly} representation that connects VLM translations with robot actions by guiding the VLM to assemble words from a semantic vocabulary.
The key idea is to first decompose the intermediate representation into a semantic vocabulary and a compositional grammar, jointly considering VLM readability and task scalability.
Then, we guide the VLM translation by constraining the VLM to compose the vocabulary words in a semantically rich and logically coherent manner, transforming the code generation into a semantics-guided assembly process.

Specifically, 
\textit{regarding VLM-comprehensibility}, the vocabulary and grammar are designed in a semantically meaningful manner for VLM-friendly translation,~\eg,
%For example:
\begin{lstlisting}[language=Python, caption=]
get_centroid(object_part_name:Str)->Point
move_cost(pt1:Point,pt2:Point,offset:List)->Cost
\end{lstlisting}
\textit{Regarding Action-generalizability}, the vocabulary provides versatile building blocks for representing diverse actions,~\eg, to translate ``cutting the carrot with a grasped knife'' into our representation, VLM can effectively produce
\begin{lstlisting}[language=Python, caption=]
perpendicular_cost(get_axis("carrot"),get_axis("knife blade"))
+move_cost(get_centroid("knife"), get_centroid("knife blade"), offset=[0,0,0.1])
\end{lstlisting}
This compositional design enables the use of a compact vocabulary that can be flexibly assembled to cater a broad range of unseen tasks.
Also, to allow precise interaction with specific object parts,~\eg, a teapot handle or a bottle cap, we build an open-vocabulary image database, consisting of paired RGB images and segmentation masks, for fine-grained object part segmentation localization.
Furthermore, we formulate a few-shot-learning-based paradigm in the retrieval-augmented generation (RAG) pipeline for accurate, instruction-specified part segmentation.

To evaluate the effectiveness of \ourmethodnospace, we design two quantitative metrics, namely {\em action-generalizability\/} and {\em VLM-comprehensibility\/}.  
The experimental results manifest that \ourmethod is able to achieve strong performance on both metrics.
Also, it largely improves real-world robot manipulation performance by 15\% over prior SOTA approaches, showing its superior robustness for practical usage.

Below, we summarize the main contributions:
\begin{itemize}
\item We design the new Semantic Assembly representation (\ourmethodnospace) 
%Composable Semantic Representation (CSR) 
for VLM-based robot manipulation, bridging the gap between VLM reasoning and robot action execution.
\item 
We introduce an RAG-based few-shot learning segmentation pipeline, enabling efficient, fine-grained, object-part level open-vocabulary segmentation.
\item 
To our best knowledge, this is the first study on analyzing the intermediate representation for VLM-based robot manipulation by formulating two new metrics, VLM-comprehensibility and action-generalizability. Extensive experiments demonstrate the compelling performance of \ourmethod over prior SOTAs by 15\%. 
\end{itemize}

\begin{figure*}[!t]
    \centering
    \includegraphics[width=\linewidth]{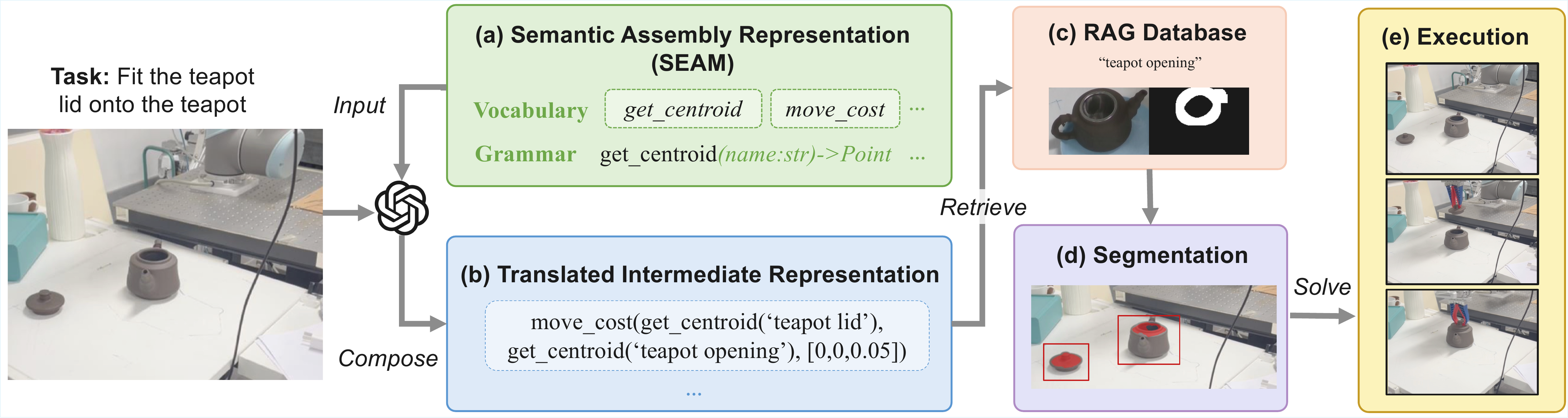}
    \vspace*{-3mm}
    \caption{Overall pipeline of our method. Given the current observation and the task instruction, our method first generates the (a) Semantic Assembly Representation (\ourmethodnospace)
    %generation CSR 
    with designed vocabulary and grammar, and then (b) translated into an intermediate representation. Next, we retrieve the corresponding support images and support masks from (c) the Retrieval Augmented Generation (RAG) Database and (d) segment the target object parts in the scene. 
    %harness our RAG-based open-vocabulary segmentation method to retrieve corresponding support images and support masks to locate necessary object parts. 
    Finally, we solve the gripper's trajectories for (e) robotic execution.
    }
    \label{fig:method_pipeline}
    \vspace{-6pt}
\end{figure*}

\section{Related Work}
\label{sec: related work}

%%%%%%%%%%%%%%%%%%%%%%%%%%%%%%%%%
\textbf{Foundation Model for Robot Manipulation}: 
Recent research on robot manipulation often relies on foundation models.
In general, there are two main types:
(i) VLA-based methods~\cite{brohan2022rt, brohan2023rt, kim24openvla, black2024pi_0, lv2025f1, gao2025vla, cheang2024gr, bu2025agibot, black2025pi_} finetune a pretrained VLM end-to-end with action data to generate robot actions in a see-and-act manner. Despite its high-frequency control and simple design, it requires massive labeled data for training.
(ii) VLM-based methods~\cite{liu2024moka, wang2024dart, liang2023code, ahn2022can, song2023llm, huang2023instruct2act, shentu2024llms,liu2024relep,mandi2024roco} bridge human instructions and robot actions via an intermediate language.
Some works design high-level semantic APIs as the intermediate representation and harness a VLM to decompose tasks into a sequence of API calls. %\phil{what is the meaning of them? it is unclear. Use some other words}.
Some other works harness the spatial relation of objects as the intermediate representation,~\eg, keypoints~\cite{huang2024rekep}, manipulation primitive~\cite{pan2025omnimanip}, and task-relevant 3D component~\cite{huang2024copa}.
Another series of works~\cite{duan2024manipulate, mu2024robocodex, huang2023voxposer,tang2025geomanip} employ a VLM to directly generate code to derive the robot actions.

%%%%%%%%%%%%%%%%%%%%%%%%%%%%%%%%%

\vspace{2mm}
\noindent
\textbf{Open-Vocabulary Object Detection and 
%Open-Vocabulary 
Part Segmentation.} 
Open-vocabulary object detection and open-vocabulary part
segmentation are challenging problems, due to training data scarcity and semantic complexity.
Regarding open-vocabulary object detection, some works~\cite{gu2021open, cheng2024yolo, du2022learning, kuo2022f, wu2023aligning, zhong2022regionclip} attempt to align text embedding 
%and 
with image embedding in the latent space, whereas others~\cite{kuo2022f, wu2023cora, wu2023clipself} finetune a pretrained backbone on a large amount of visual data and harness the extracted visual-semantic features.
Some more recent works~\cite{bai2025qwen2,li2025lmm,fu2025llmdet} propose to finetune a VLM with object detection data to harness the VLM's ability on spatial and semantic understanding.

Regarding open-vocabulary part segmentation, some works~\cite{tang2025geomanip, zhang2023personalize, zhou2024point, yuan2024open} suggest various mechanisms to prompt the class-agnostic segmentation model,~\eg, SAM~\cite{kirillov2023segment}, to directly predict the mask.
Some others~\cite{lai2024lisa,zou2023generalized, wang2024llm,yuan2025sa2va,wei2023ov} finetune a VLM with segmentation data to align the mask generation with the pretrained semantics.
However, satisfactory performance is hard to achieve in segmenting object parts for robot manipulation.

%%%%%%%%%%%%%%%%%%%%%%%%%%%%%%%%%

\vspace{2mm}
\noindent
\textbf{Automata Theory for Robot Manipulation.}
We observe a strong connection between the design of the intermediate representation and early works in modeling and controlling robots with automata theory.
System modeling with discrete, event-driven logic is a well-established field. 
The first work dates back to~\cite{ramadge1987supervisory}, in which the automata theory~\cite{rabin1959finite} 
is utilized to model an events system. 
Later, hybrid automaton~\cite{alur1991hybrid, cassandras2007introduction, henzinger1996theory,lygeros2003dynamical,hristu2005handbook} combines a Finite Automaton with differential equations.
%to develop hybrid control system. 
More recently,~\cite{dantam2012motion} proposes Motion Grammar to model robot manipulation with context-free grammar and discusses the completeness, correctness, and decidability for manipulation tasks. 
Notations in our work are inspired by context-free grammar. For better readability, we use the term vocabulary to collectively refer to lexicon, grammar for rules, and sacrifice rigorous formulation, without loss of generality.
\begin{figure*}[t]
    \centering
    \includegraphics[width=\linewidth]{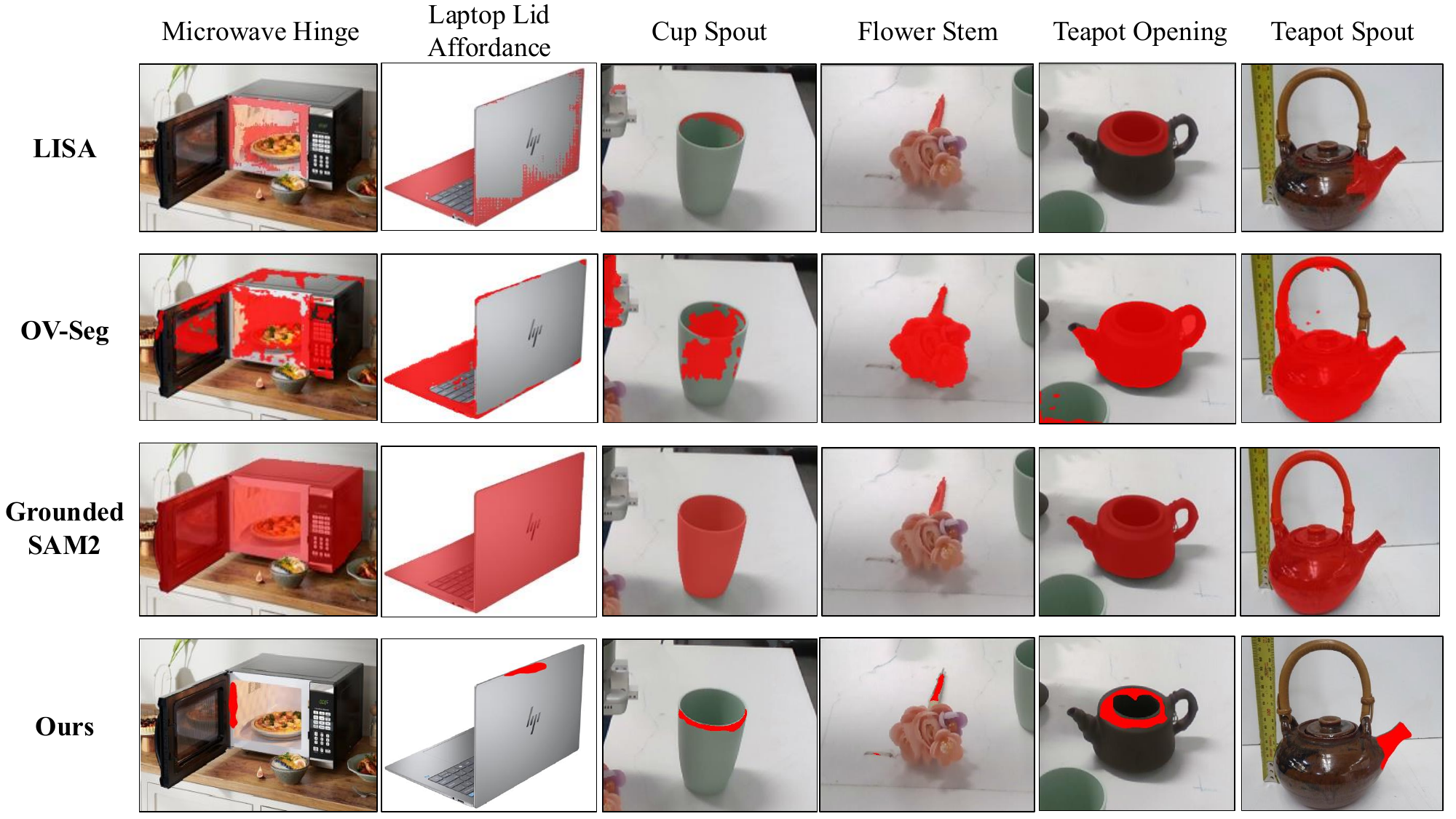}
    \caption{Qualitative performance comparisons for open vocabulary segmentation between the state-of-the-art methods and our methods on common manipulation.
    }
    \label{fig:segmentation_comp}
\end{figure*}
\label{subsec: segmentation_method}

\section{Method}
In this section, we first present the problem formulation (Section~\ref{subsec:preliminary_problem_formulation}). 
Then, we explain the relationships between SEAM and context-free grammar (Section~\ref{subsec:cfg}), elaborate on the design of SEAM (Section~\ref{subsec: representation_design}), present our novel open-vocabulary segmentation paradigm
%open-vocabulary localization to support robot manipulation.
%
(Section~\ref{subsec: segmentation_method}), and 
%
%Lastly, we 
lastly discuss how we convert the intermediate representation to robot actions (Section~\ref{subsec:traj_gen}).
The overall pipeline of our method is illustrated in Fig.~\ref{fig:method_pipeline}.

%%%%%%%%%%%%%%%%%%%%%%%%%%%%%%%%%%%%%%%%%%%%%%%%%%%%%%%%%%%%%%%%%%%%%%%%%%%%%%%%%%

\subsection{Problem Formulation}
\label{subsec:preliminary_problem_formulation}
Given a human instruction $L$ and visual input $I$, the VLM-based robot manipulation task can be formulated as a translation process:
$\mathcal{R} = \text{VLM}(L,I),$
where $\mathcal{R}$ is an intermediate representation, which can be passed into a solver to generate the robot actions, such as the gripper's 3D position $\{x,y,z\}$, 3D orientation (\eg, Euler angles $\{X,Y,Z\}$), and gripper status $O$.
We formulate $\mathcal{R}$ as a linguistic representation: $\tilde{{\mathcal{R}}}=(\mathcal{V}, \mathcal{G})$, where $\mathcal{V}$ is the vocabulary that comprises a set of skill words and $\mathcal{G}$ is the grammar that specifies the rules for their valid composition. The core challenge lies in the design of the skill words in the vocabulary and rules in the grammar, to make them amenable for VLM to generate, while being expressive to facilitate zero-shot generalization to handling diverse and unseen tasks.

\subsection{Context-Free Grammar and SEAM}
\label{subsec:cfg}
A Context-Free Grammar (CFG) consists of a finite set of symbols (words) and a finite set of recursive production rules, that together enable us to model a language and generate string patterns within the language space. Formally, it can be defined as a 4-tuple $(V, \Sigma, R, S)$, which includes a set of non-terminal symbols $V$, terminal symbols $\Sigma$, production rules $R$ that map non-terminals to sequences of terminals and non-terminals, and a start symbol $S$. CFG provides a structured yet flexible framework for defining the syntax of formal languages, enabling unambiguous parsing and interpretation of complex symbolic expressions. 

We adapt the powerful CFG framework to design \ourmethod and to create a more intuitive and semantically-grounded interface for VLM, as shown in Tab.~\ref{tab:our_method_design}. SEAM leverages the structural rigor of a CFG but is specifically designed for VLM comprehensibility. The SEAM's vocabulary $\mathcal{V}$ is parallel to a union of sets $V$, $\Sigma$, and $S$ in CFG, yet offering semantically-rich, human-readable words for modeling robot manipulation tasks. The SEAM's grammar $\mathcal{G}$ is parallel to the production rules $R$ in CFG, yet the rules are formatted in a semantically-rich and human-readable manner.

\begin{table}[t]
    \centering
    \caption{The design of \ourmethod composes of vocabulary $\mathcal{V}$ and grammar $\mathcal{G}$ listed below. 
    % \phil{TODO: the grammar part of this table is not easy to follow; if you have space in paper, provide more explanation about it in the main text}
    }
    \resizebox{\columnwidth}{!}{  
    \begin{tabularx}{0.45\textwidth}{s@{\hspace{1mm}}s|X}
       \toprule
       \multicolumn{2}{c|}{Vocabulary $\mathcal{V}$} & \multicolumn{1}{c}{Grammar $\mathcal{G}$} \\ 
       \hline
         {\small get\_axis,} \newline
         {\small get\_centroid,} \newline
         {\small get\_height,} \newline 
         {\small move\_cost,} \newline  
         {\small parallel\_cost,} \newline
         {\small get\_gripper\_pos} \newline
         {\small perpendicular\_cost,} 
         & 
         {\small rotate\_cost,} \newline
         {\small orbit\_cost,} \newline
         {\small gripper\_close,} \newline 
         {\small gripper\_open,} \newline 
          
         &
         {\small cost $\rightarrow$ cost + cost,} \newline 
         {\small $\text{get\_axis,object} \rightarrow \text{vec}$,} \newline
         {\small $\text{get\_centroid,object} \rightarrow \text{pt}$,} \newline
         {\small $\text{get\_height,object}$$\rightarrow$$\text{cost}$,} 
         {\small $\text{parallel\_cost,vec,vec}$$\rightarrow$$\text{cost}$,} \newline
         {\small $\text{move\_cost,pt,pt} \rightarrow \text{cost}$,} \newline
         {\small $\text{pt} \rightarrow \text{pt} \pm \text{pt}$}
        \\
        \bottomrule
    \end{tabularx}
    }
%    \caption{Design of vocabulary $\mathcal{V}$ and grammar $\mathcal{G}$ for our \ourmethodnospace.}
    \label{tab:our_method_design}
    \vspace{-10pt}
\end{table}

\subsection{SEAM Representation Design}
\label{subsec: representation_design}
We elaborate on the design of our representation here. Specifically, we will design our vocabulary $\mathcal{V}$ and grammar $\mathcal{G}$. Our ultimate goal is to balance the VLM-comprehensibility and action-generalizability. In order to achieve this, we stick to the following principles:
\begin{itemize}
\vspace*{1mm}
\item \textbf{VLM-Readability:} Vocabulary words in $\mathcal{V}$ are grounded in the VLM's semantic space for accurate comprehension.
\vspace*{-3.5mm}
\item \textbf{Proper Abstraction:} The representation abstracts away implementation details, exposing only the parameters essential for task planning and generalization.
\vspace*{1mm}
\item \textbf{Conciseness:} Semantics are conveyed through a minimal sequence of symbols, using high-level words to maximize information density.
\vspace*{1mm}
\item \textbf{Reliability:} A type system within $\mathcal{G}$ constrains the composition of $\mathcal{V}$, ensuring syntactically valid and semantically sound output from the VLM.
\vspace*{1mm}
\item \textbf{Proper Minimalism:} The core words are kept small and intuitive to reduce the VLM's learning burden and prevent unexpected behaviors.
\vspace*{1mm}
\item \textbf{Composability:} The representation is inherently modular, supporting extensibility through the seamless integration of new primitives and rules.
\end{itemize}

%\vspace*{2mm}
% \newpage
Our design helps to achieve good VLM-readability, as all the words in the vocabulary are rich in semantics and close to human natural language. We achieve good abstraction since all the words in the vocabulary abstract away the repetitive and tedious underline implementation (e.g., using principal components analysis to calculate the axis in ``get\_axis''), leaving only necessary parameters exposed (e.g., the points as input in ``get\_axis''). Our design aims to make vocabulary orthogonal, that is, to minimize semantic overlap among elements, thereby achieving conciseness. The grammar rules in $\mathcal{G}$ define the data type for vocabulary assembly, which constrains the VLM to assemble the vocabulary according to the grammar rules and produce more reliable outputs. By keeping our vocabulary and grammar rules minimum, the VLM can be more focused in 
%has less confusion to 
distinguishing words and makes the output more predictable. Though minimal, our words in the vocabulary can assemble a wide range of actions via the grammar rules, showcasing our design's strong extensibility.

%%%%%%%%%%%%%%%%%%%%%%%%%% RAG-base Open-vocabulary Segmentation %%%%%%%%%%%%%%%%%%%%%%%%%%%%%%%%%%%%%%%%%%%%%%%%
\subsection{RAG-based Few-shot Open-vocabulary Segmentation}

Grounding target object parts (\eg, teapot opening, flower stem) on words such as \textit{``get\_centroid''} and \textit{``get\_axis''} in $\mathcal{V}$ requires a precise and fine-grained open-vocabulary method. 
As Fig.~\ref{fig:segmentation_comp} shows, existing state-of-the-art segmentation models struggle to achieve satisfactory performance.
For example, OV-Seg~\cite{wei2024ov} and Grounded SAM2~\cite{ren2024grounded} tend to segment the whole objects/parts rather than associated parts for robot interaction, whereas LISA~\cite{lai2024lisa} falls short of segmenting the hinge, opening, and affordance.
To support robot manipulation with fine-grained object part segmentation, we build a database $\mathcal{D} = \{ (\mathcal{K}_i, \mathcal{P}_i) \}_{i=1}^N$, where 
$\mathcal{K}$ is a set of key phrases that describe specific object parts,~\eg, the key phrases of the object part ``cup opening'' can be \{cup opening, cup rim, cup edge\}; and $\mathcal{P}$ is a set of support pairs defined as $\mathcal{P} = \{{ (I{_{j}}^S, M_{j}^S) } \}_{j=1}^n$, where $I_j^S$ is the $j$-th support image containing the object part and $M_j^S$ is the associated binary support mask that highlights the object part in $I_j^S$. 
Given a query image $I_Q$ and a natural language instruction \textit{desc}, we retrieve all support pairs $\mathcal{P}_i$ with the matched key phrase $\mathcal{K}_i$. Specifically, we calculate the Levenshtein distance~\cite{lcvenshtcin1966binary} between \textit{desc} and each key phrase in $\mathcal{K}_i$, and target $\mathcal{K}_i$ contains a key phrase with the least Levenshtein distance to \textit{desc}. The details of our database for RAG-based open-vocabulary segmentation are provided in the Appendix.

In detail, during the generation phase, our target is to generate query mask $M_Q$ based on the retrieved $\mathcal{P}_i$.  Here, we adopt a few-shot segmentation network as $\text{Mapper}$~\cite{tang2025overcoming}, 
% which is an improved version of~\cite{shi2022dense} that performs better when number of support images are larger than 10. 
% $\text{Mapper}$ 
which essentially maps the support masks $M_S$ to the query mask $M_Q$, according to the support feature and query feature similarity calculated as the attention score. 
% We refer readers to~\cite{tang2025overcoming} for the detailed design.

%%%%%%%%%%%%%%%%%%%%%%%%%%%%% Trajectory Generation %%%%%%%%%%%%%%%%%%%%%%%%%%%%%%%%%%%%%
\subsection{Trajectory Generation}
\label{subsec:traj_gen}
To convert our intermediate representation to robot actions, we solve for the gripper end pose to minimize the overall costs evaluated by our intermediate representation. 
Specifically, the SEAM representation is Python-executable, and the execution result is a numerical cost that captures point cloud ${P}$ matches the SEAM representation.  Next, we locate which object parts are moving along with the gripper by identifying whether they belong to the grasped object. The point clouds of these moving object parts are denoted as $P^m$, while stationary object parts are represented as $P^s$.

Since the gripper is rigidly attached to the point cloud of the moving parts $\mathcal{P}^m$, they should share the same transformation. Thus, solving for the gripper's target rotation $\mathbf{R} \in SO(3)$ and translation $\mathbf{t} \in \mathbb{R}^3$ reduces to the following optimization problem:
\begin{equation}
    \begin{aligned}
        \min_{\mathbf{R}, \mathbf{t}}\ 
        & \text{language}\left(P^s \cup \left(\mathbf{R}\mathbf{R}_0^{-1} (P^m -\mathbf{t}_0) + \mathbf{t}\right)\right) \\
        & + \alpha \|\mathbf{t} - \mathbf{t}_0\|_2 + \beta \|\text{euler}(\mathbf{R}\mathbf{R}_0^{-1})\|_1,
    \end{aligned}
    \label{eq:optimize}
\end{equation}
where $\mathbf{R}_0$ and $\mathbf{t}_0$ represent the gripper's initial pose, $\text{euler}(\cdot)$ extracts the Euler angles. The last two terms are the regularization terms with $\alpha$, $\beta$ as the two weights to encourage the gripper moves with the least transition and rotation, respectively.
\
\begin{figure}[!t]
    \centering
    \includegraphics[width=\linewidth]{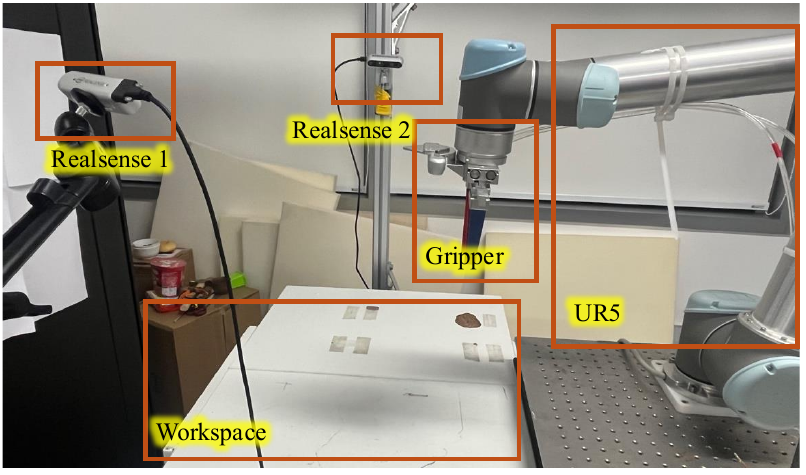}
    \caption{Our system setup includes a robotic module (A UR5 robot equipped with a gripper) and a visual perception module (two realsense D435 cameras on opposite side of the workspace).}
    \label{fig:setup}
    \vspace{-10pt}
\end{figure}

\begin{figure*}[t]
    \centering
    \includegraphics[width=0.95\linewidth]{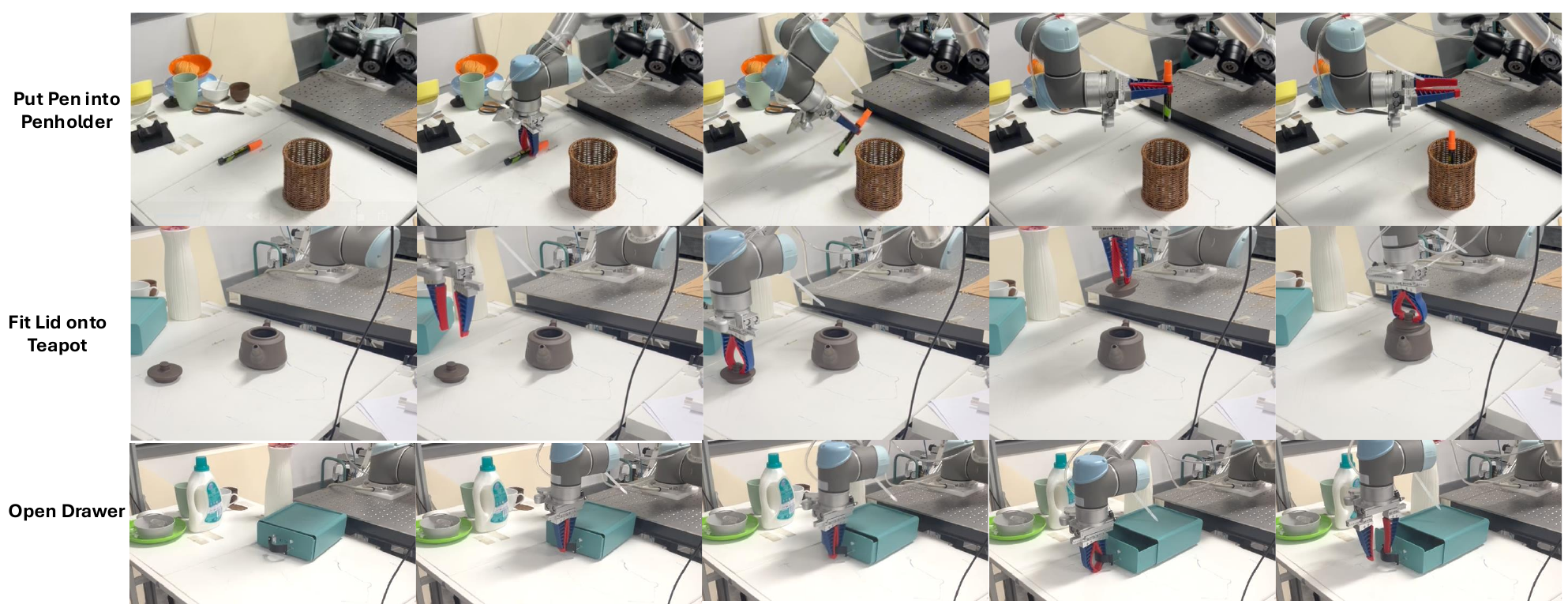}
    \caption{Execution sequences of the three tasks in the real-world environment. The sequences of all the tasks can be found in the Appendix.}
    \label{fig:main_exp_seq}
\end{figure*}
\section{Experiment}
%%%%%%%%%%%%%%%%%%%%%%%%%%%%%%%%%%%%%%%%%%%%%%%%%%%%%%%%%%%%%%%%%%%%%%%%%%%%

\subsection{Experimental Setup}
\textbf{Hardware}\quad 
%As illustrated in 
Fig.~\ref{fig:setup} depicts our hardware setup, in which our robotic platform is built using the UR5 industrial robot with a gripper as its end-effector.
To obtain visual perception, we set up a dual-camera module by arranging two Intel RealSense D435 depth-sensing cameras, strategically positioned on the opposite sides of the workspace.
Doing so enables us to obtain a comprehensive stereo coverage of the operational area for more robust 3D scene perception.
\newline
%%%%%%%%%%%%%%%%%%%%%%%%%%%%%%%%% Deployment %%%%%%%%%%%%%%%%%%%%%%%%%%%%%%%%%
\textbf{Implementation Details}\quad 
%For
%
We deployed Qwen3-VL-30B-22A~\cite{yang2025qwen3} on an A100 GPU as the VLM in our method.
Regarding the RAG-based open-vocabulary segmentation, we adopt the Swin-B transformer~\cite{liu2021swin} for feature extraction and the trained matcher from~\cite{tang2025overcoming}.

\subsection{Reference Prompt for Intermediate Representation Generation}
Similar to our comparison methods~\cite{huang2023voxposer, huang2024copa,huang2024rekep,pan2025omnimanip}, to improve the robustness of intermediate representation generation, we build a prompt for the VLM models to refer to. The prompt consists of intermediate representation for some ``atomic action''. As an example, for the action ``move something relative to something'', the intermediate representation is ``move\_cost(`[something]', `[something]', offset)''. These templates provide strong hint to prevent the VLM from hallucination or generating incomplete representation. The complete prompt for ``atomic action'' and the corresponding representation are listed in the Appendix.

\subsection{Evaluation Tasks and Performance Measures}
We developed a suite of eight distinct activities to test robotic manipulation performance under realistic conditions. The benchmark includes six tasks requiring interaction with rigid objects (such as pouring beverages) and six additional tasks involving articulated mechanisms (like drawer opening). This diverse selection incorporates various object types to examine generalization capacity and adaptation in challenging settings. Each method underwent 10 repeated trials per task with success rates documented as the primary metric. To prevent evaluation bias, we randomized object configurations between trials.

\subsection{Comparison Methods}
Our method is evaluated against four established methods:

\myparagraph{VoxPoser}~\cite{huang2023voxposer} synthesizes robot motion trajectories by constructing 3D value maps through large language and vision-language models, demonstrating strong zero-shot reasoning and continuous feedback capabilities;

\myparagraph{CoPa}~\cite{huang2024copa} incorporates component-level spatial limitations integrated with visual-language understanding to facilitate manipulation without vocabulary restrictions;

\myparagraph{ReKep}~\cite{huang2024rekep} implements relationship-based keypoint constraints and uses multi-level optimization to translate natural language commands into immediate action sequences.

\myparagraph{OmniManip}~\cite{pan2025omnimanip} uses the manipulation-aware primitives to construct spatial constraints for action solving.

%%%%%%%%%%%%%%%%%%%%%%%%%%%%%%%%% Main Results %%%%%%%%%%%%%%%%%%%%%%%%%%%%%%%%%
\subsection{Main Results}
We conduct experiments on eight tasks and the success rates are illustrated in Tab.~\ref{tab:main_result}. Following~\cite{pan2025omnimanip}, we have two variants with close-loop and open-loop planning. The objects poses and positions are randomly initialized.
% under the following two conditions: 1. All essential object parts are clearly captured by the two RGBD cameras in the environment. 2. The objects are within the robot's workspace. 
We can conclude that our method achieves consistent better zero-shot generalizability across wide ranges of tasks with different objects poses and positions, surpassing the OmniManip by $15\%$ in the average success rates. The execution sequences of tasks are illustrated in Fig.~\ref{fig:main_exp_seq}. 
The complete execution sequences for the eight tasks are listed in the Appendix. 
% We also refer the readers to the supplementary for the execution video for the 8 tasks. 
Next, we analyze how to achieve the performance with our two key designs: (i) \ourmethod, and (ii) RAG-based open-vocabulary segmentation.

\begin{table*}[t]
\centering
\caption{Performance comparison across different methods}
\label{tab:results}
\begin{tabular}{lccccccc}
\toprule
\multirow{4}{*}{Tasks} & \multicolumn{7}{c}{Methods} \\
\cmidrule{2-8}
 & VoxPoser & CoPa & ReKep & \multicolumn{2}{c}{OmniManip} &\multicolumn{2}{c}{\ourmethod (ours)}\\
\cmidrule{5-8}
 & & & & Closed-loop & Open-loop &Closed-loop& Open-loop\\
\midrule
% Pour tea & 0/10 & 1/10 & 3/10 & 7/10 & 6/10 & & \\
% Insert flower into vase & 0/10 & 4/10 & 2/10 & 6/10 & 4/10 & & \\
Insert the pen in holder & 0/10 & 4/10 & 3/10 & 7/10 & 5/10 & \textbf{8/10} & 6/10\\
Recycle the battery & 6/10 & 5/10 & 7/10 & 8/10 & 6/10 &\textbf{8/10} & 6/10 \\
Pick up cup/bowl onto the dish & 3/10 & 2/10 & 9/10 & 8/10 & 7/10 &\textbf{9/10} & 5/10 \\
Fit the lid onto the teapot & 0/10 & 2/10 & 3/10 & 5/10 & 3/10&\textbf{7/10} & \textbf{7/10} \\
Open the drawer & 1/10 & 4/10 & - & 6/10 & 4/10& \textbf{8/10} & 5/10  \\
Close the drawer & 3/10 & 3/10 & - & 8/10 & 6/10&\textbf{9/10} & 5/10  \\
% Hammer the button & 0/10 & 3/10 & - & 4/10 & 2/10 & & \\
Press the red button & 0/10 & 3/10 & - & 7/10 & 6/10&\textbf{10/10} & 9/10 \\
% Close the lid of the laptop & 4/10 & 3/10 & - & 6/10 & 4/10& &  \\
Open the jar & 2/10 & 0/10 & - & 6/10 & 5/10& \textbf{8/10} & 8/10  \\
\textbf{Total} & 18.6\% & 28.8\% & - & 68.8\% & 52.5\%& \textbf{83.8}\%& 63.8\% \\
\bottomrule
\label{tab:main_result}
\end{tabular}
\vspace{-20pt}
\end{table*}

\begin{figure}
    \centering
    \includegraphics[width=\linewidth]{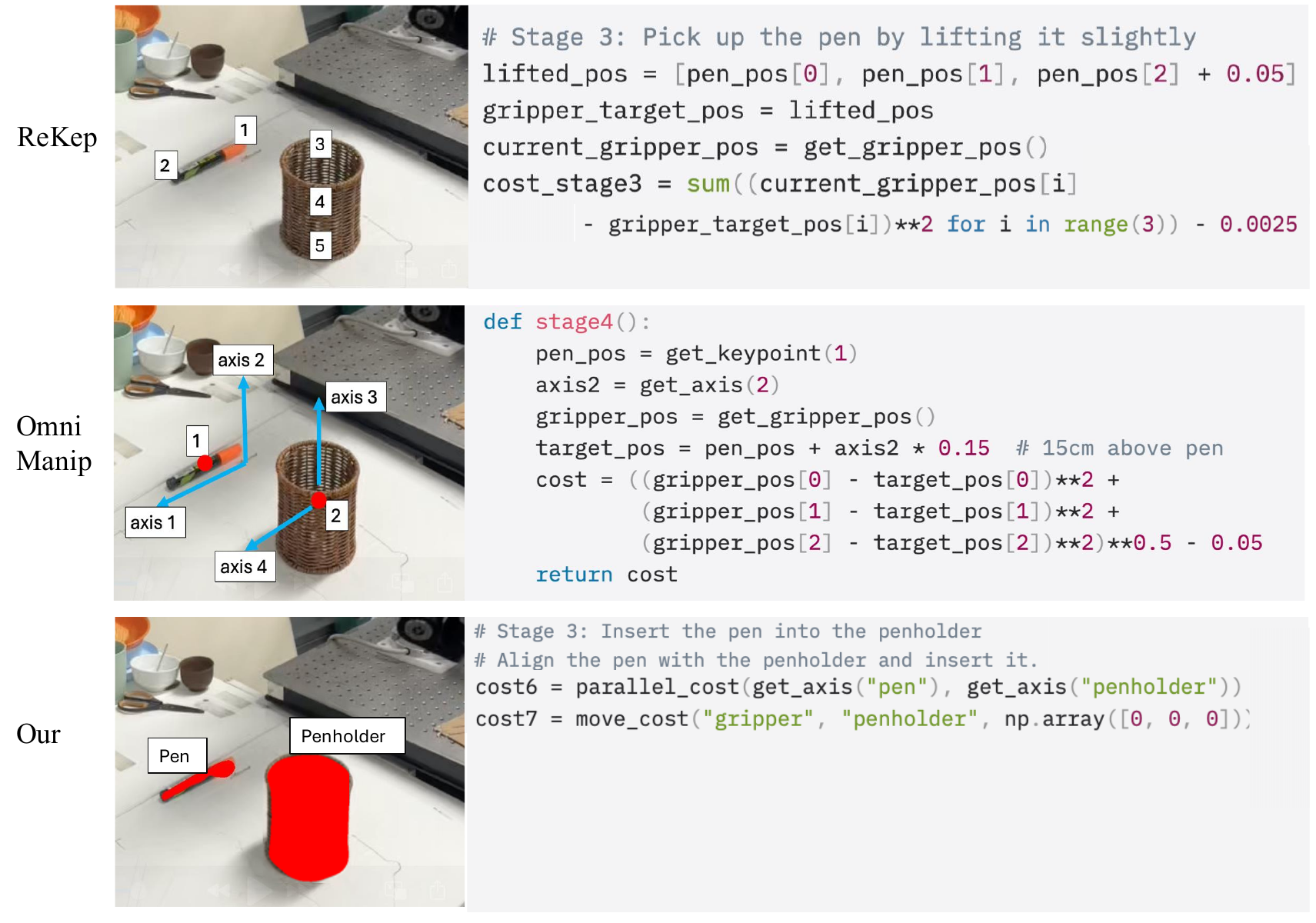}
    \caption{Comparisons among the intermediate representation generated by state-of-the-art methods and our methods for the task ``put the pen into the penholder''.}
    \label{fig:main_result_analysis}
    \vspace{-15pt}
\end{figure}

% \begin{figure}
%     \centering
%     \includegraphics[width=\linewidth]{figures/tradeoff_plot.pdf}
%     \caption{Statistics on each method's action-generalizability (action-gen) and VLM-comprehensibility (VLM-comp).
%     \phil{Interact2act is so good? Please correct this plot against Figure 9. Also, make it smaller? Mention the publication year and venue [XXX, 20XX] of other people's works?}
%     }
%     \label{fig:scattar}
% \end{figure}

\textbf{Semantic Assembly Representation} 
Our \ourmethod allows VLM to comprehend and compose the intermediate representation. As illustrated in Fig.~\ref{fig:main_result_analysis}, for the task ``put the pen into the penholder'', in the stage ``move the pen above the penholder'',  ReKep~\cite{huang2024rekep} struggles to align the pen's axis with the penholder's axis, requires the VLM to solely use ``get\_keypoint'' to select the correct keypoints and calculate the axis from the labeled image. OmniManip~\cite{pan2025omnimanip} also struggles to align the pen's axis with the penholder's axis, which requires the VLM to select from the image the correct axis and compose the correct representation. In contrast, our \ourmethod abstracts away the complications of direction alignment with semantic-rich vocabulary and grammar, so it can easily generate the correct representation to align the pen and the penholder's axis. 
% A more systematic and quantitative experiment is conducted in~\ref{subsec:preliminary_study}

\begin{figure}
    \centering
    \includegraphics[width=\linewidth]{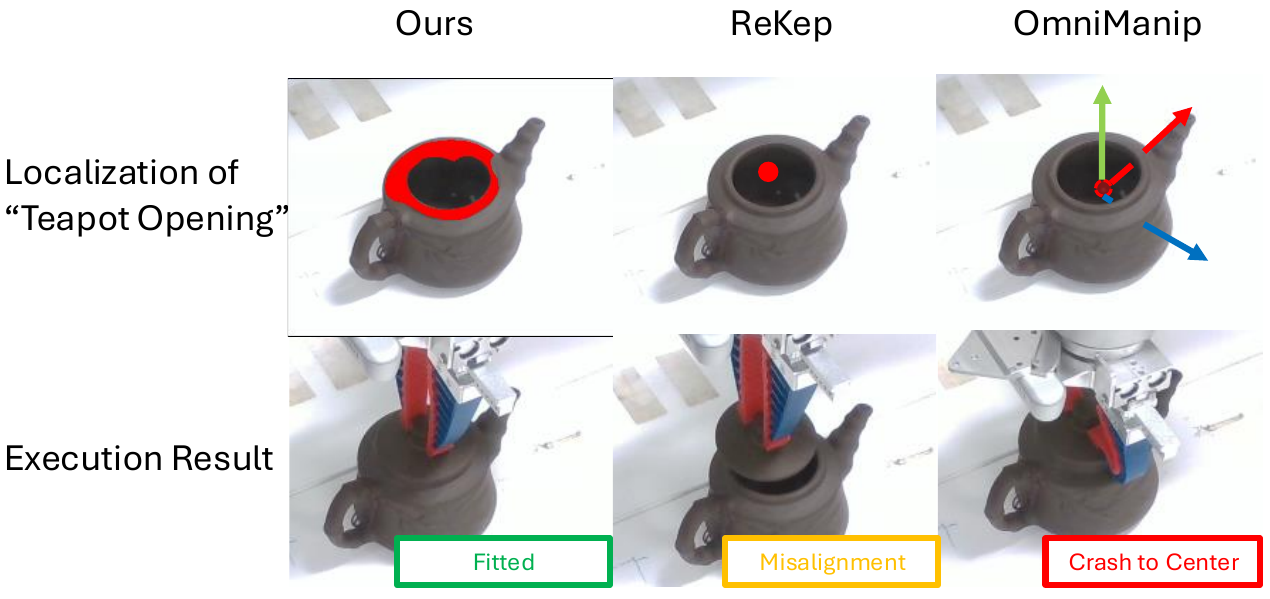}
    \caption{Comparisons among open-vocabulary localization for ``teapot opening''and their corresponding execution results in the task ``fit the teapot lid on the teapot''.}
    \label{fig:segm_case_study}
\end{figure}
%
%%%%%%%%%%%%%%%%%%%%%%%%%%%%%%%%% RAG-based Open-Vocabulary Segmentation %%%%%%%%%%%%%%%%%%%%%%%%%%%%%%%%%
\textbf{RAG-based Open-Vocabulary Segmentation} The RAG-based open-vocabulary segmentation pipeline enables us to flexibly locate any object part essential for manipulation. We qualitatively compare the performance of our methods with existing SOTA methods in Fig.~\ref{fig:segmentation_comp}. More segmentation comparison results can be found in the Appendix. In our real-world experiment, our segmentation method to a large extent helps improve the success rates on tasks requiring careful alignment. For example, in the task ``fit the lid on the teapot'', we successfully locate the rim of the teapot opening, which provides us with an accurate location to move the lid down, as illustrated in Fig.~\ref{fig:segm_case_study}. Other methods can only locate the teapot center~\cite{pan2025omnimanip, liang2023code, huang2023voxposer} or the interior point inside the teapot~\cite{huang2024rekep} as illustrated in Fig.~\ref{fig:segm_case_study}, which may lead to misalignment or crushing the lid into the teapot's center. Furthermore, we deploy all the methods on an A6000 GPU to compare time elapse. The results in Tab.~\ref{tab:time_comp_segm} demonstrate the efficiency of our method.

\begin{table}[t]
    \centering
    \small
    \setlength{\tabcolsep}{5.4pt}
    \caption{Comparisons of time elapse between our method and the state-of-the-art methods.}
    \begin{tabular}{c|c@{\hspace{2mm}}c@{\hspace{2mm}}c@{\hspace{2mm}}c} 
    \toprule
    Model & LISA & OV-seg & Grounded SAM & SEAM (ours) \\
    \midrule
     Time (sec.)  & 0.9 & 10.2 & 0.88 & \textbf{0.6} \\
    \bottomrule
    \end{tabular} 
    \label{tab:time_comp_segm} 
    \vspace{-15pt}
\end{table}

%%%%%%%%%%%%%%%%%%%%%%%%%%%%%%%%% RAG-based Open-Vocabulary Segmentation %%%%%%%%%%%%%%%%%%%%%%%%%%%%%%%%%
\subsection{Quantitative Study on Action-Generalizability and VLM-Comprehensibility}
\label{subsec:preliminary_study}
In this section, we compare the performance of existing mainstream VLM-based robot manipulation methods on the design of intermediate representation, and prove our Semantic Assembly Representation (\ourmethodnospace) is better. 

\myparagraph{Evaluation Metrics.}
\label{subsec:preliminary_metrics}
To evaluate the performance of intermediate representation, we focus on two metrics: (i) \textbf{Action-generalizability}: How well the intermediate representation can generalize to unseen manipulation tasks. (ii) \textbf{VLM-comprehensibility}: With the spatial reasoning and semantic understanding of the existing VLMs, whether the intermediate representation can be correctly generated given the action instruction. 
% Formally, action-generalizability is associated with the minimum number of $\mathcal{V}$ that is required to  translate all $I$ into $L$.
% VLM-comprehensibility measures given human instruction $I$, whether $\text{VLM}$ generates action such that after executed, the environment semantically fits the human instruction.
Formally, action-generalizability (AG) is defined as:
$$
\text{AG} = 1 - \frac{|\mathcal{V}|}{T},
$$
where $|\mathcal{V}|$ denotes the number of unique vocabulary operations required to translate all human instructions $I$, 
and $T$ is the total number of manipulation tasks. A higher action-generalizability indicates that fewer vocabulary operations are needed to represent all tasks, implying stronger generalization ability.
VLM-comprehensibility (VC) measures the proportion of tasks that a VLM can successfully complete given human instructions, which is formulated as:
$$
\text{VC} = \frac{N_{\text{succ}}}{T},
$$
where $N_{\text{succ}}$ is the number of successfully executed tasks.
% \begin{table}[t]
%     \centering
%     \begin{tabularx}{0.5\textwidth}{l|X|X}\toprule
%        Method  & action-gen\newline ($|\mathcal{V}|$/\# tasks) & VLM-comp\newline (\# tasks/\# correct)\\ 
%        \hline
%         ReKep     & 33/6 & 9/33  \\
%         \hline
%         OmniManip & 33/8 & 15/33 \\
%         \hline
%         Instruct2act & 33/24 & 26/33 \\
%         \hline
%         Ours & 33/12 & 22/33 \\
%         \bottomrule
%     \end{tabularx}
%     \caption{Statistics on each method's action-generalizability (action-gen) and VLM-comprehensibility (VLM-comp)}
%     \label{tab:analytic_exp_result}
% \end{table}

\begin{figure}
    \centering
    \includegraphics[width=0.9\linewidth]{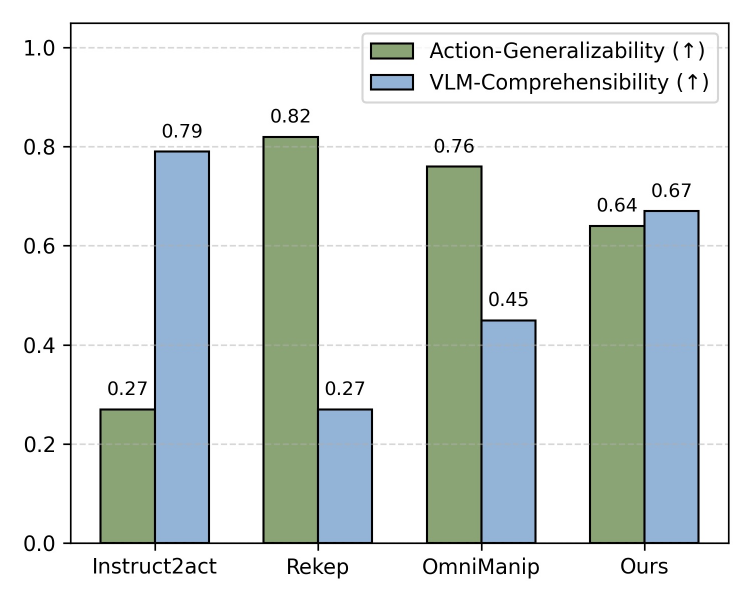}
    \caption{Statistics on each method's action-generalizability and VLM-comprehensibility.}
    \label{fig:analytic_exp_result}
\end{figure}

\myparagraph{Baselines.}
\label{subsec:preliminary_exp}
We compare \ourmethod on action-generalizability and VLM-comprehensibility with one high-level representation method: \textbf{Instruct2Act}~\cite{huang2023instruct2act} uses predefined action APIs to construct intermediate representation, and two low-level representation methods: (i) \textbf{ReKep} ~\cite{huang2024rekep} adopts keypoints for intermediate representation. (ii) \textbf{Omnimanip}~\cite{pan2025omnimanip} uses intermediate representation based on axis and keypoint primitives. We also summarize the vocabulary $\mathcal{V}$ for each of the methods, partly listed in Tab.~\ref{tab:method_setting}. The complete vocabulary $\mathcal{V}$ and grammar is $\mathcal{G}$ listed in Appendix.
% \begin{table*}[t]
%     \centering
%     \begin{tabularx}{\textwidth}{t|b|s}\toprule
%        Method  & $\mathcal{V}$ & $\mathcal{G}$ \\ 
%        \hline
%         ReKep     & get\_keypoint, gripper\_close, gripper\_open, move\_to, get\_gripper\_pos, get\_gripper\_pose, $\mathcal{V}_{\text{Python}}$ & $\text{cost}\rightarrow \text{cost}+\text{cost}$;\newline
%         $\text{cost}\rightarrow$cost\_fns,kpts;\newline
%         kpts$\rightarrow$get\_keypoint;
%          $\mathcal{G}_{\text{Python}}$\\
%         \hline
%         OmniManip & get\_keypoint, get\_axis,get\_keypoint, gripper\_close, gripper\_open, move\_to, get\_gripper\_pos, get\_gripper\_pose & $\text{cost}\rightarrow \text{cost}+\text{cost}$;\newline 
%         $\text{cost}\rightarrow$ distance constraint, p, p;\newline
%         p$\rightarrow$ get\_keypoint;
%         p$\rightarrow$ get\_axis
%         \\
%         \hline
%         Instruct2act & gripper\_close, gripper\_open, get\_gripper\_pos, get\_gripper\_pose, find, pick, place, pick\_place, insert, push, press, move\_above, move\_perpendicular, move\_parallel, flip, scoop, stir, tap, screw\_rotation, controlled\_pour, route\_around, pull\_apart, fold, straighten, $\mathcal{V}_{\text{Python}}$ &$\text{action}\rightarrow \text{action} + \text{action}$; \newline
%         $\text{action}\rightarrow \text{v, segm}$;\newline
%         $\text{segm}\rightarrow \text{find, obj}$;
%         $\mathcal{G}_{\text{Python}}$\\
%       \bottomrule
%     \end{tabularx}
    
%     \caption{$\mathcal{V}$ and $\mathcal{G}$ for each method}
%     \label{tab:method_setting}
% \end{table*}
\begin{table}[t]
    \centering
    \caption{$\mathcal{V}$ and $\mathcal{G}$ for each method.
    % \phil{Do you really need this in main paper? How about putting it in supp.?  When writing research papers, try not to create problems that reviewers may attack.}
    }
    \begin{tabularx}{\linewidth}{p{2.5cm}|X}\toprule
       Method  & Vocabulary $\mathcal{V}$ \\ 
       \hline
        ReKep     & get\_keypoint, move\_to, $\mathcal{V}_{\text{Python}}$ \\
        \hline
        OmniManip & get\_keypoint, get\_axis, move\_to, \\
        \hline
        Instruct2act & find, pick, place, pick\_place, insert, push,
        press, flip, scoop, stir, tap, fold, straighten, move\_above, \newline
        move\_perpendicular,route\_around,
        move\_parallel,  screw\_rotation, \newline
        controlled\_pour,   $\mathcal{V}_{\text{Python}}$ \\
      \bottomrule
    \end{tabularx}
    \label{tab:method_setting}
    \vspace{-10pt}
\end{table}

% $\textbf{V}_{\text{Python}}$ and $\textbf{G}_{\text{Python}}$ represents the Python vocabulary(keywords) and Python grammar(Python arithmetic, Python expression), respectively.
\myparagraph{Task.} We randomly generate 33 manipulation tasks on single-arm, non-tactile and non-force-feedback. For each method, we prompt Qwen3-VL~\cite{yang2025qwen3} with the corresponding $\mathcal{V}$ and $\mathcal{G}$ to generate intermediate representation. Since it's laborious-infeasible to validate the success rate with a real or virtual environment for all the tasks, we employ DeepSeek~\cite{guo2025deepseek} to help evaluate whether the intermediate representation can reasonably produce actions that complete the tasks successfully. The implementation details can be found in the Appendix.

\myparagraph{Experiment Results.}
\label{subsec:preiminary_exp_result}
The action generalizability and the VLM-comprehensibility results are demonstrated in Fig.~\ref{fig:analytic_exp_result}
From the experimental results, we can clearly validate: (i) There is a trade-off between the action-generalizability and VLM-comprehensibility. Previous state-of-the-art methods' intermediate representations are either too high-level (high VLM-comprehensibility, low action-generalizability)~\cite{liang2023code} or too low-level(low VLM-comprehensibility, high action-generalizability)~\cite{huang2024rekep, pan2025omnimanip}. (ii) Our \ourmethod balances the VLM-comprehensibility and action-generalizability.

\section{Conclusion}

In this work, we observe that there is a tradeoff between the VLM-comprehensibility and the action-generalizability among the intermediate representations chosen by various VLM-based robot manipulation methods. To address this, we create Semantic Assembly Representation (\ourmethodnospace) by designing our vocabulary and grammar stick to our principles. To localize the object part for reprensetation evaluation, we design a RAG-based open-vocabulary segmentation method which can efficiently and effectively segment fine-grained object parts. Additionally, We introduce two metrics to demonstrate the scalability and comprehensibility of VLMs for our \ourmethod. Our analytical experiment results show that our method can generate intermediate representation robustly and achieve state-of-the-art performance in the real-world experiments.

\clearpage
{
    \small
    \bibliographystyle{ieeenat_fullname}
    \bibliography{main}
}


\clearpage
\appendix
\addtocontents{toc}{\protect\setcounter{tocdepth}{2}}
\onecolumn
\begin{center}
\section*{\huge Appendix}
\end{center}

\setcounter{figure}{0}
\setcounter{table}{0}

% \addtocontents{toc}{\protect\setcounter{tocdepth}{-10}}
% \setcounter{tocdepth}{2} 

\tableofcontents
\newpage

\section{Real-world Execution Sequences for the Eight Tasks}
\label{appx:complete_seq}
\begin{figure}[H]
    \centering
    \includegraphics[width=0.95\linewidth]{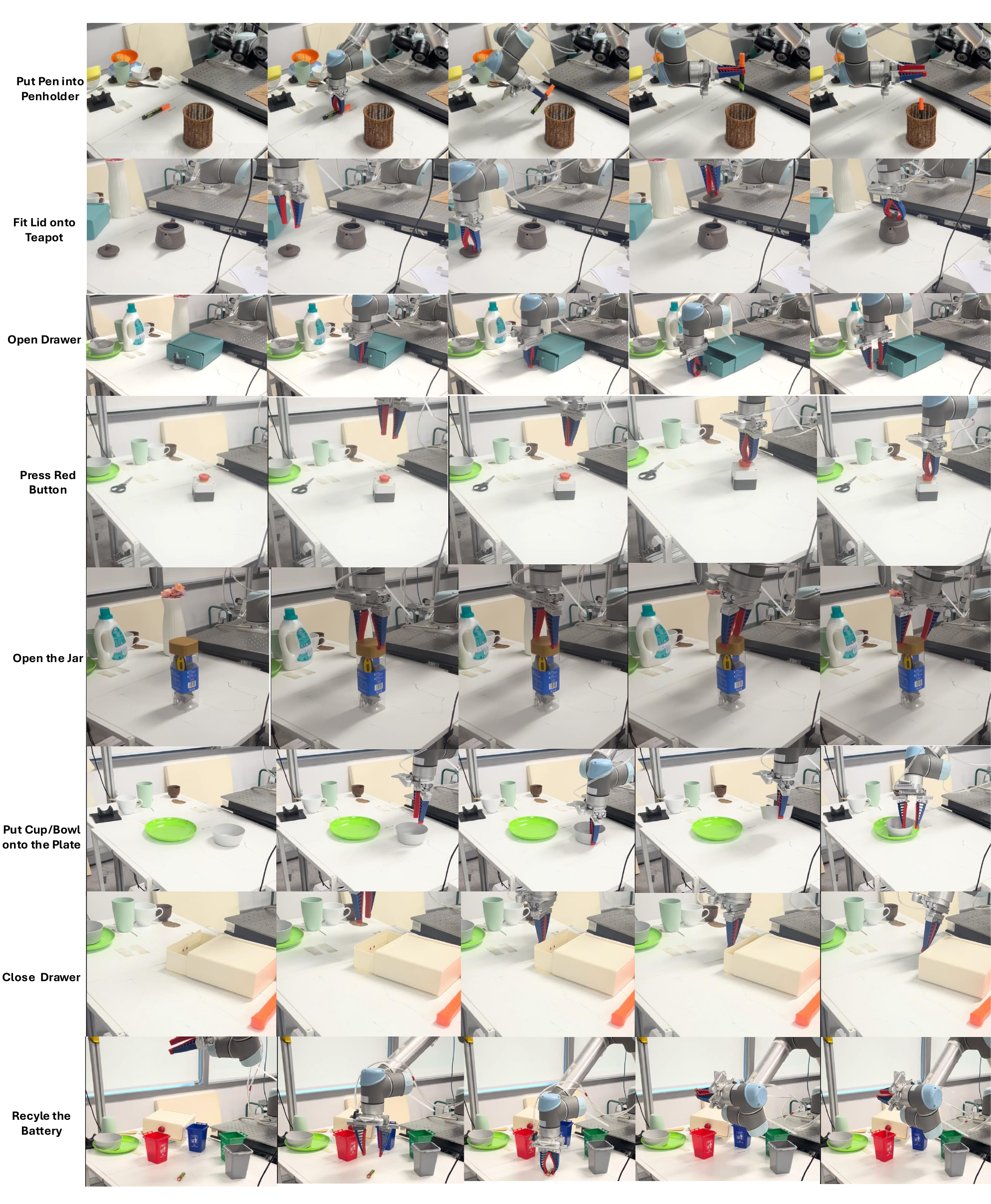}
    \caption{The sequences of execution for the eight real-world tasks.}
    \label{fig:seq_appx}
\end{figure}

In this section, we demonstrate the execution sequence of the eight real-world tasks in~\ref{fig:seq_appx}.

From the execution sequence we can see our method can manage diverse tasks including non-prehensile manipulation (e.g., press red button, open the jar), pick-and-place (e.g., put cup/bowl onto the plate), articulation object manipulation (e.g., open/close drawer), manipulation with alignment (e.g., put pen into penholder), manipulation with precise localization (e.g. fit lid onto teapot).

\section{Robust Experiment Settings on Scenes and Task}
\begin{figure}[H]
    \centering
    \includegraphics[width=0.9\linewidth]{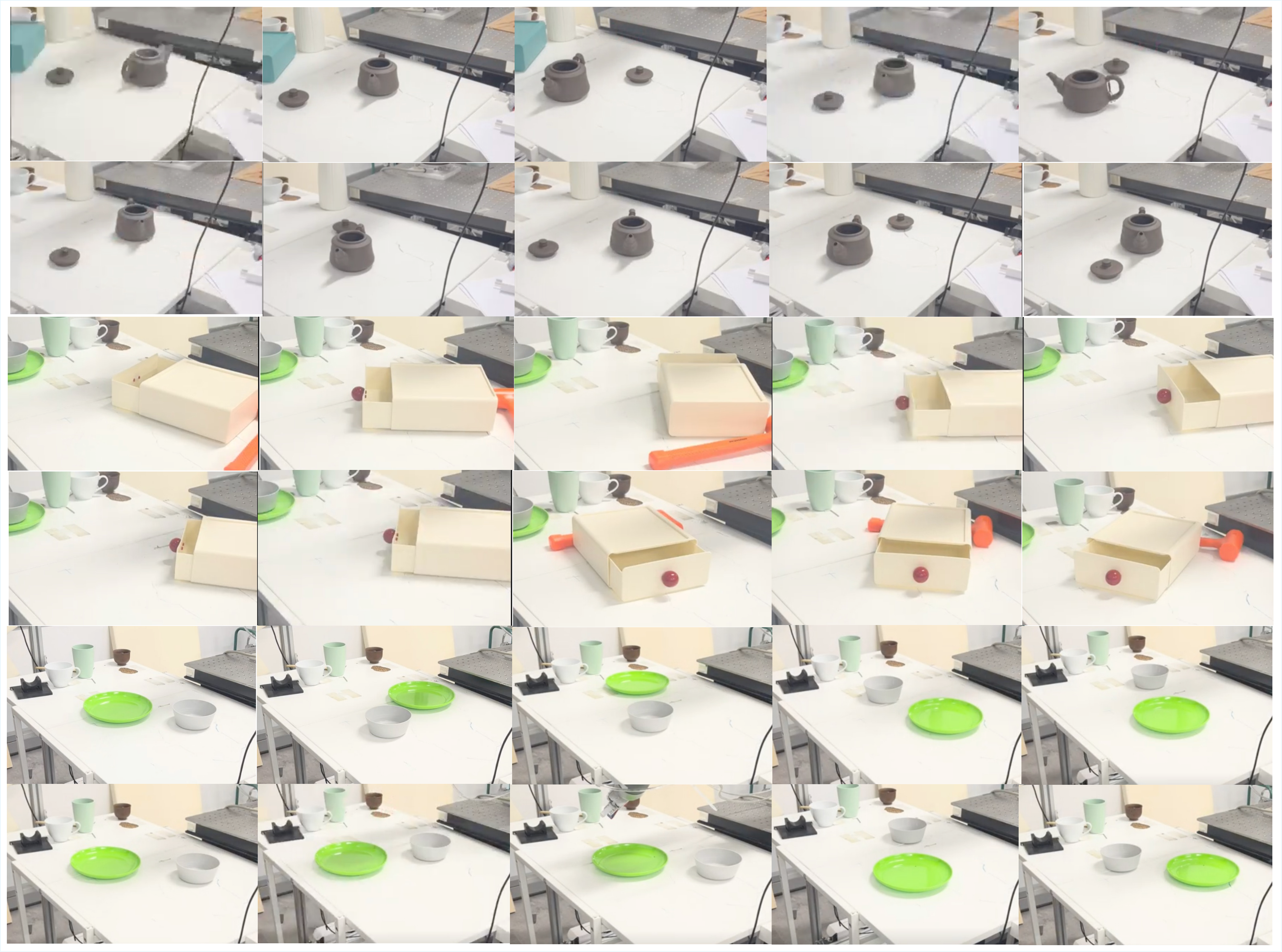}
    \caption{The initial scene settings.}
    \label{fig:init_appx}
\end{figure}

We illustrate the ten scene settings for three tasks in Fig~\ref{fig:init_appx}. For each of the tasks, we fully randomize the object initial positions, object initial orientations, and background object placements under the robot workspace and camera view volume.

\section{Tasks List in Quantitative Study on Action-Generalizability and VLM-Comprehensibility}
The 33 tasks we employed in the preliminary experiments are listed as follows:
\begin{enumerate}
\item Sort the Red Cube: Pick the red cube from a group and place it inside the red circle.
\item Bin the Blue Cylinder: Pick the blue cylinder and drop it into the bin marked with a blue square.
\item Stack Cube on Cube: Pick the green cube and stack it on top of the yellow cube.
\item Move the Soda Can: Pick the soda can from the left table and place it on the right table.
\item Fill the Tray: Pick the AA battery and place it into the empty slot in the plastic tray.
\item Insert the USB Drive: Pick the USB drive from the table and insert it into the laptop's USB port.
\item Assemble the LEGO: Pick the 2x4 LEGO brick and attach it to the red baseplate, connecting it to two other bricks.
\item Place the Ring: Pick the wooden ring and place it onto the vertical post.
\item Put the Lid on the Jar: Pick the plastic jar lid and place it on top of the jar
\item Hang the Key: Pick the key and hang it on the keyhook by its hole.
Non-Prehensile Manipulation
\item Push the Dice: Push the white dice across the table until it crosses the black line.
\item Flip the Pancake: Use the spatula to flip the pancake in the frying pan.
\item Close the Drawer: Push the kitchen drawer closed using the flat of the gripper.
\item Press the Doorbell: Press the round, lit doorbell button on the wall.
\item Align the Block: Push the wooden block until it is flush against the corner of the table.
\item Scoop the Rice: Use the metal spoon to scoop rice from the pot into the bowl.
\item Stir the Soup: Use the spoon to stir the liquid in the pot three times clockwise.
\item Hammer the Nail: Use the toy hammer to tap the nail until its head is flush with the board.
\item Screw in the Lightbulb: Pick the lightbulb and screw it into the empty lamp socket.
\item Pour the Water: Pick the pitcher and pour water into the empty glass until it is half-full.
\item Uncoil the Rope: Manipulate the coiled rope until it forms a straight line from start to end.
\item Fold the Washcloth: Fold the small, square washcloth in half.
\item Open the Bag: Use two grippers to pull the handles of the plastic bag apart.
\item Drape the Towel: Drape the hand towel over the horizontal bar.
\item Route the Cable: Route the USB cable around the two posts in an S-shape.
\item Grasp the Marble: Pick the glass marble from a flat surface.
\item Grasp the Coin: Pick the single coin from the table.
\item Re-grip the Screwdriver: Pick the screwdriver by its handle, then place it down and re-grip it by its shaft.
\item Pick the Book: Pick the paperback book from the shelf by its spine.
\item Hook the Mug: Hook a gripper finger through the handle of the coffee mug and lift it.
\item Place the T-Block: Pick the T-shaped block and insert it into the matching T-shaped slot on the board.
\item Assemble the Stack: Pick the large square block and place it on the table, then place the medium block on it, and finally the small block on top.
\item Plug in the Lamp: Pick the power plug from the floor and insert it into the wall outlet.
\end{enumerate}

These tasks cover a wide range of tasks in including basic prehensile manipulation, precise manipulation, tool use, deformable object manipulation, non-prehensile manipulation, and assembly, which provide a comprehensive evaluations on VLM's robustness to generate intermediate representation and reflects the VLM's comprehensibility.

\section*{}

\newpage
\section{Complete Vocabulary and Grammar Table for Different Comparison Methods}
To the best of our knowledge, this is the first study on analyzing the intermediate representation for VLM-based robot manipulation. Following the VLM-based robot manipulation works, we reformulate and summarize their prompt designs into vocabulary and grammar. The complete vocabulary and grammar settings for each method are illustrated in Tab.~\ref{tab:vocab_grammar_setting_complete}.
\label{appx:vocab_grammar}
\begin{table*}[h]
    \centering
    \begin{tabularx}{\textwidth}{l|X|X}\toprule
       Method  & Vocabulary $\mathcal{V}$ & Grammar $\mathcal{G}$ \\ 
       \hline
        ReKep    & get\_keypoint, gripper\_close, gripper\_open, move\_to, get\_gripper\_pos, get\_gripper\_pose, $\mathcal{V}_{\text{Python}}$& $\text{cost}\rightarrow \text{cost}+\text{cost}$,\newline  $\text{cost}\rightarrow$ cost\_fns, kpts,\newline
        kpts$\rightarrow$kpts,keypoint,\newline
        kpts$\rightarrow$get\_keypoint,\newline
        kpts$\rightarrow$get\_end\_effector,\newline
        $\mathcal{G}_{\text{Python}}$\\
        \hline
        OmniManip & gripper\_close, gripper\_open, move\_to, \newline get\_gripper\_pos, get\_gripper\_pose, get\_keypoint, get\_axis & $\text{cost}\rightarrow \text{cost}+\text{cost}$,\newline $\text{cost}\rightarrow$ angular constraint, p, p,\newline 
        $S\rightarrow$ distance constraint, p, p,\newline
        p$\rightarrow$ get\_keypoint,\newline
        p$\rightarrow$ get\_axis
        \\
        \hline
        Instruct2act & gripper\_close, gripper\_open, get\_gripper\_pos, \newline get\_gripper\_pose, find, pick, place, pick\_place, \newline insert, push, press, move\_above, move\_parallel, move\_perpendicular, flip, scoop, stir, tap, \newline screw\_rotation, controlled\_pour, route\_around, \newline pull\_apart, fold, straighten, $\mathcal{V}_{\text{Python}}$ &$\text{action}\rightarrow \text{action} + \text{action}$, \newline $\text{action}\rightarrow \text{v, segm}$, \newline 
        $\text{segm}\rightarrow \text{find, obj}$, \newline  get\_gripper\_pos $\rightarrow$ action,\newline
        find $\rightarrow$ action,\newline
        pick $\rightarrow$ action,\newline
        place $\rightarrow$ action,\newline
        pick\_place $\rightarrow$ action,\newline
        insert $\rightarrow$ action,\newline
        push $\rightarrow$ action,\newline
        press $\rightarrow$ action,\newline
        move\_above $\rightarrow$ action,\newline
        move\_parallel $\rightarrow$ action,\newline
        ... \newline
        $\mathcal{G}_{\text{Python}}$\\
        \bottomrule
    \end{tabularx}
    
    \caption{Vocabulary $\mathcal{V}$ and Grammar $\mathcal{G}$ for each method.}
    \label{tab:vocab_grammar_setting_complete}
\end{table*}

From the vocabulary and grammar table, we can observe that works with high-level intermediate representation (Instruct2act) needs extensive vocabulary to cover our designed 33 tasks. Meanwhile, the vocabulary of the works with low-level intermedate representation(RekKep, OmniManip) is abstract and poor in VLM-comprehensibility.

% \newpage

\section{``Atomic Action'' and their associated Intermediate Representation}
\label{appx:rag_db_for_lang_gen}

\begin{enumerate}
    \item move something to something with an offset
    \begin{lstlisting}
        move_cost('<source object part>', centroid('<target object part>') + np.array([<x, y, z>])) + parallel_cost(get_axis('<source object part>'), vector) + upright_cost(up_part='<up object part of the same source object>', down_part='<down object part of the same source object>')

## 0. You can you can use function get_height('<object part>'), get_width('<object part>'), get_length('<object part>') to get the dimension of the object part to calculate for the offset
## 1. [x, y, z] is the EXTRA between the <target object part> and the <source object part>
    ## 1.0 If move above the target, [<x, y, z>] = [0, 0, some positive margin (> 0.1) + get_cheight('<target object part>')]
    ## 2.0 If move to contact the target from front, [<x, y, z>] = [0, 0, 0]
## 2 (optional) Keep the parallel cost if <source object part> part need to be aligned with some vector direction after the move
    ## 2.0 if <source object part> aligned with the target object part, vector = get_axis('<target object part>')
    ## 2.1 if <source object part> aligned vertically, vector = [0, 0, -1]
## 3 (optional) If the <source object> needs to be upright after moved, keep upright_cost. down_part and up_part MUST be the same object!
## 4 All object part should follow the format "object part of the object"
    \end{lstlisting}
    \item pick something up or put something down when grasped
    \begin{lstlisting}
        move_cost_with_offset('<source object part>', offset=[<0, 0, z>])

## 0. You can use function get_height('<object part>'), get_width('<object part>'), get_length('<object part>') to get the dimension of the object part to calculate for the offset
## 1. If move up / lift, z= get_height('<object part>') + a positive margin
## 2. If move down / place,  z= - get_height('<object part>') + a negative margin
    \end{lstlisting}
    \item press something after aligned
    \begin{lstlisting}
        gripper_close_first_cost() + move_cost('gripper', '<target object part>')
    \end{lstlisting}
    \item pull to open something after grasped
    \begin{lstlisting}
        move_cost('gripper', centroid_last('gripper') + direction_of(start='<object you want to pull away from>', end='gripper') * <offset_distance in meters>)

## fill in the <object you want to pull away from> accordingly.
## <offset_distance in meters> > 0.1
    \end{lstlisting}
    \item push to close something after grasped
    \begin{lstlisting}
        gripper_open_cost()

## directly use gripper_open_cost() to open the gripper and release the grasped item
    \end{lstlisting}
    \item release something only
    \begin{lstlisting}
        gripper_open_cost()

## directly use gripper_open_cost() to open the gripper and release the grasped item
    \end{lstlisting}
\end{enumerate}

\section{The RAG Database for Open-vocabulary Part Segmentation}
\label{appx:rag_db_for_segm}
Below, we visualize example cases from our RAG database used for open-vocabulary part segmentation.
The six visualized cases include the microwave hinge, laptop lid affordance, cup spout, flower stem, teapot opening, and teapot spout. 
Each case has one or more object part segmentation examples, along with the words used to prompt the segmentation.

\begin{figure}[H]
    \centering
    \includegraphics[width=\linewidth]{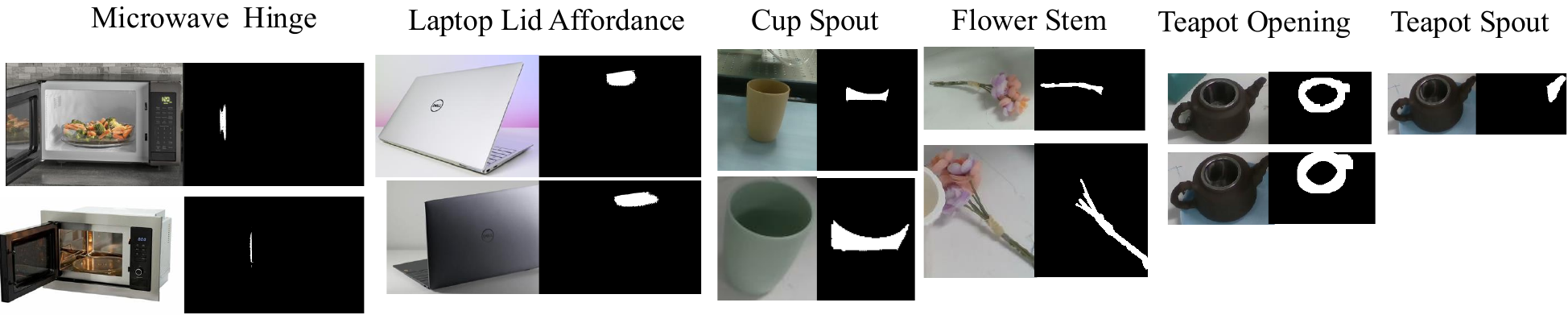}
    \caption{Example cases from the RAG database for open-vocabulary part segmentation.
    }
    \label{fig:placeholder}
\end{figure}
\label{appx:seq}

\newpage
\section{More Results on Open-vocabulary Segmentation}
In this section, we present the open-vocabulary segmentation results from our method and three state-of-the-art techniques: LISA, OV-Seg, and Grounded SAM2. 
We focus on five cases: button, doorbell, hammer cap, key ring, and pen cap, with results highlighted in blue. 
Our method demonstrates a higher precision in identifying critical object parts essential for robotic manipulation.
For instance, it precisely locates the button to be pressed, the specific area on the doorbell, the cap of the hammer used in striking, the key ring for hanging keys, and the cap of the pen.
In comparison, Grounded SAM2 often segments entire objects instead of key parts and may misidentify background elements.
OV-Seg produces messy masks and typically identifies whole objects. 
Although LISA can estimate the locations of target parts, its masks can be messy and may occasionally encompass the entire object.

\label{appx:more_segm_comp}
\begin{figure}[H]
    \centering
    \includegraphics[width=\linewidth]{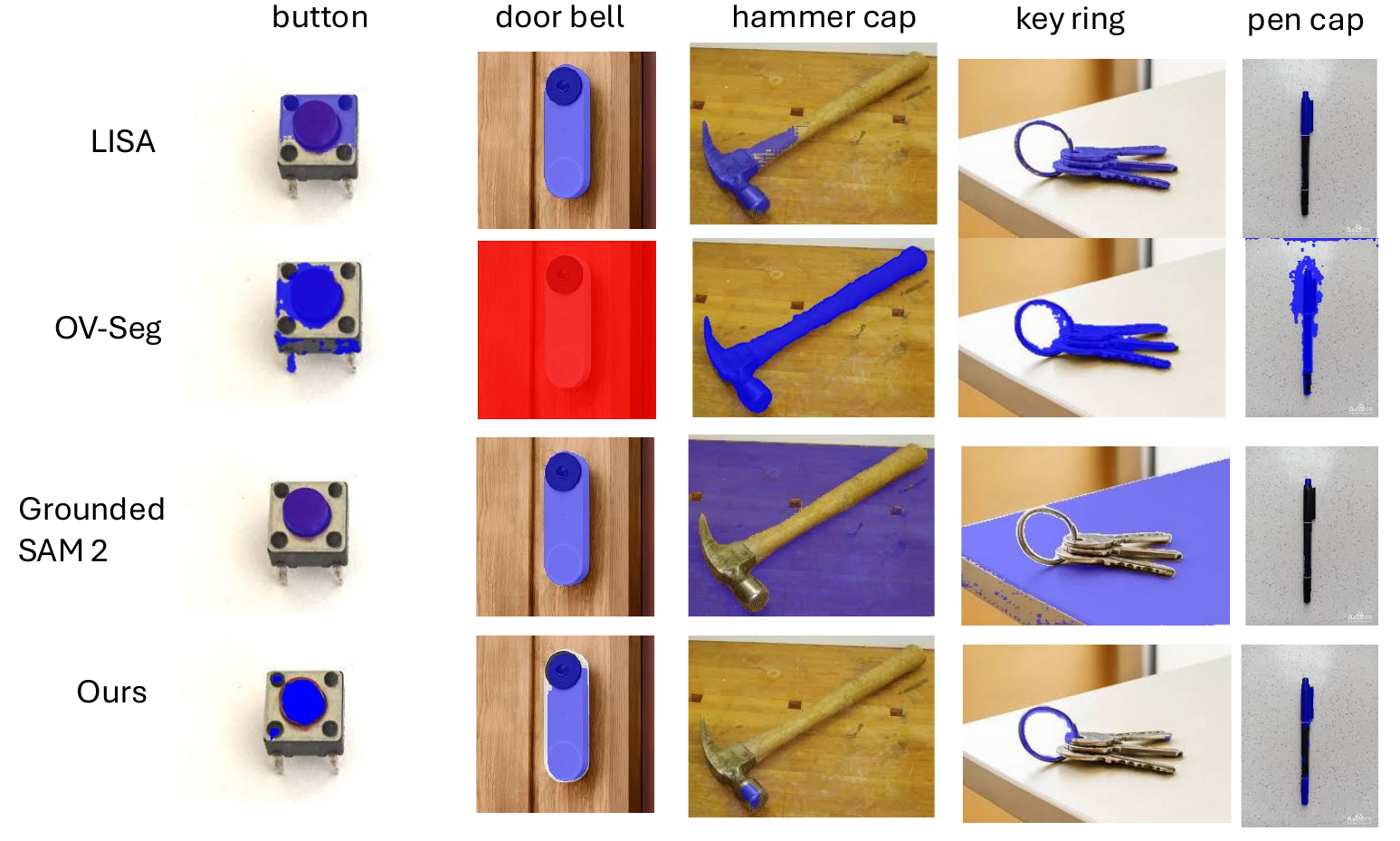}
    \caption{More open-vocabulary segmentation comparisons with other SOTA methods and our method.}
    \label{fig:more_segm_comp}
\end{figure}
% \clearpage

\section{RAG Database for Open-vocabulary Part Segmentation in Real-world Experiments.}
\label{appx:rag_db_for_segm_rw}

Below, we provide some examples from our RAG database for open-vocabulary part segmentation in real-world experiments. 
We include 15 cases: button, bowl rim, bin, battery, bowl, basket, teapot opening, teapot spout, pen cap, plate, teapot lid, pen, drawer handle, drawer, and cap. 
Each case features a text prompt along with one or multiple segmentation results.

\begin{figure}[H]
    \centering
    \includegraphics[width= \linewidth]{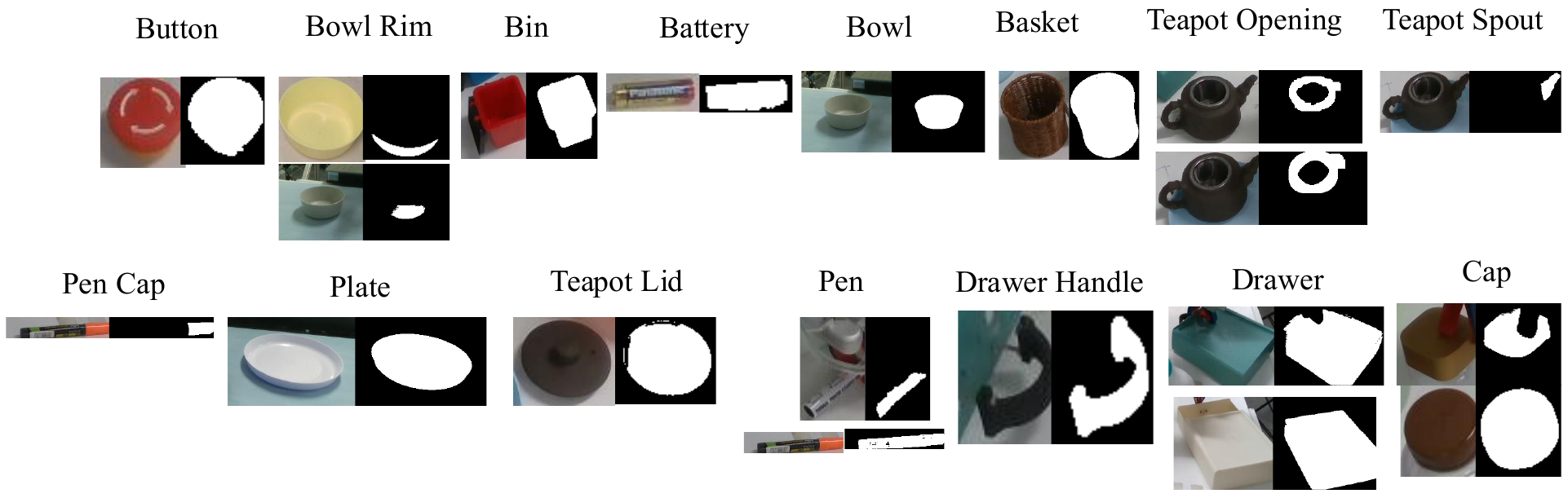}
    \caption{Some example cases from the RAG database for open-vocabulary segmentation in the real-world experiments.
    }
    \label{fig:placeholder}
\end{figure}

% \clearpage
\section{Prompt and the Generated Outputs Used in the Analytical Experiments}
\label{appx: analytical_exp_output}

\subsection{OmniManip}
\textbf{Prompt}: 
\begin{lstlisting}
    For each of the task, decompose it into stages and write me the Python intermediate code for each stage. Each code consists of Python basic arithmetic and choose extra function ONLY from the vocabulary: $[$ get\_keypoint, get\_gripper\_pose, get\_gripper\_xyz, get\_axis, gripper\_close, gripper\_open, move\_to$]$. get\_keypoint (obj:str) function returns the 3D coordinate (xyz) of the point of an object, and get\_axis(obj:str) function returns the three principle axes(pitch, yaw, or roll axis) of obj. get\_gripper\_pose() returns the XYZ Euler angle of the gripper, get\_gripper\_pos() return the xyz coordinate of the gripper. move\_to(x,y,z,X,Y,Z) move the gripper to location(xyz) and pose(XYZ). Your should organize your output in Latex with enumerate and lstlisting.
\end{lstlisting}

\myparagraph{\textbf{Generated Language By Qwen3-VL}}:
\begin{enumerate}
\item \textbf{Sort the Red Cube:} Pick the red cube from a group and place it inside the red circle.
\begin{lstlisting}
# Stage 1: Move to red cube
cube_pos = get_keypoint("red_cube")
move_to(cube_pos[0], cube_pos[1], cube_pos[2] + 0.1, 0, 0, 0)
gripper_open()
move_to(cube_pos[0], cube_pos[1], cube_pos[2], 0, 0, 0)
gripper_close()

# Stage 2: Move to red circle
circle_pos = get_keypoint("red_circle")
move_to(circle_pos[0], circle_pos[1], circle_pos[2] + 0.1, 0, 0, 0)
move_to(circle_pos[0], circle_pos[1], circle_pos[2], 0, 0, 0)
gripper_open()
\end{lstlisting}

\item \textbf{Bin the Blue Cylinder:} Pick the blue cylinder and drop it into the bin marked with a blue square.
\begin{lstlisting}
# Stage 1: Move to blue cylinder
cyl_pos = get_keypoint("blue_cylinder")
move_to(cyl_pos[0], cyl_pos[1], cyl_pos[2] + 0.1, 0, 0, 0)
gripper_open()
move_to(cyl_pos[0], cyl_pos[1], cyl_pos[2], 0, 0, 0)
gripper_close()

# Stage 2: Move to bin
bin_pos = get_keypoint("blue_bin")
move_to(bin_pos[0], bin_pos[1], bin_pos[2] + 0.1, 0, 0, 0)
move_to(bin_pos[0], bin_pos[1], bin_pos[2], 0, 0, 0)
gripper_open()
\end{lstlisting}

\item \textbf{Stack Cube on Cube:} Pick the green cube and stack it on top of the yellow cube.
\begin{lstlisting}
# Stage 1: Move to green cube
green_pos = get_keypoint("green_cube")
move_to(green_pos[0], green_pos[1], green_pos[2] + 0.1, 0, 0, 0)
gripper_open()
move_to(green_pos[0], green_pos[1], green_pos[2], 0, 0, 0)
gripper_close()

# Stage 2: Move above yellow cube
yellow_pos = get_keypoint("yellow_cube")
move_to(yellow_pos[0], yellow_pos[1], yellow_pos[2] + 0.2, 0, 0, 0)
move_to(yellow_pos[0], yellow_pos[1], yellow_pos[2] + 0.05, 0, 0, 0)
gripper_open()
\end{lstlisting}

\item \textbf{Move the Soda Can:} Pick the soda can from the left table and place it on the right table.
\begin{lstlisting}
# Stage 1: Move to soda can
can_pos = get_keypoint("soda_can")
move_to(can_pos[0], can_pos[1], can_pos[2] + 0.1, 0, 0, 0)
gripper_open()
move_to(can_pos[0], can_pos[1], can_pos[2], 0, 0, 0)
gripper_close()

# Stage 2: Move to right table
right_table_pos = get_keypoint("right_table")
move_to(right_table_pos[0], right_table_pos[1], right_table_pos[2] + 0.1, 0, 0, 0)
move_to(right_table_pos[0], right_table_pos[1], right_table_pos[2], 0, 0, 0)
gripper_open()
\end{lstlisting}

\item \textbf{Fill the Tray:} Pick the AA battery and place it into the empty slot in the plastic tray.
\begin{lstlisting}
# Stage 1: Move to battery
battery_pos = get_keypoint("AA_battery")
move_to(battery_pos[0], battery_pos[1], battery_pos[2] + 0.1, 0, 0, 0)
gripper_open()
move_to(battery_pos[0], battery_pos[1], battery_pos[2], 0, 0, 0)
gripper_close()

# Stage 2: Move to tray slot
slot_pos = get_keypoint("tray_slot")
move_to(slot_pos[0], slot_pos[1], slot_pos[2] + 0.1, 0, 0, 0)
move_to(slot_pos[0], slot_pos[1], slot_pos[2], 0, 0, 0)
gripper_open()
\end{lstlisting}

\item \textbf{Insert the USB Drive:} Pick the USB drive from the table and insert it into the laptop’s USB port.
\begin{lstlisting}
# Stage 1: Move to USB drive
usb_pos = get_keypoint("USB_drive")
move_to(usb_pos[0], usb_pos[1], usb_pos[2] + 0.1, 0, 0, 0)
gripper_open()
move_to(usb_pos[0], usb_pos[1], usb_pos[2], 0, 0, 0)
gripper_close()

# Stage 2: Move to USB port
port_pos = get_keypoint("laptop_USB_port")
move_to(port_pos[0], port_pos[1], port_pos[2] + 0.05, 0, 0, 0)
move_to(port_pos[0], port_pos[1], port_pos[2], 0, 0, 0)
gripper_open()
\end{lstlisting}

\item \textbf{Assemble the LEGO:} Pick the 2x4 LEGO brick and attach it to the red baseplate, connecting it to two other bricks.
\begin{lstlisting}
# Stage 1: Move to Lego brick
lego_pos = get_keypoint("2x4_Lego_brick")
move_to(lego_pos[0], lego_pos[1], lego_pos[2] + 0.1, 0, 0, 0)
gripper_open()
move_to(lego_pos[0], lego_pos[1], lego_pos[2], 0, 0, 0)
gripper_close()

# Stage 2: Move to baseplate connection point
baseplate_pos = get_keypoint("red_baseplate_connection")
move_to(baseplate_pos[0], baseplate_pos[1], baseplate_pos[2] + 0.05, 0, 0, 0)
move_to(baseplate_pos[0], baseplate_pos[1], baseplate_pos[2], 0, 0, 0)
gripper_open()
\end{lstlisting}

\item \textbf{Place the Ring:} Pick the wooden ring and place it onto the vertical post.
\begin{lstlisting}
# Stage 1: Move to ring
ring_pos = get_keypoint("wooden_ring")
move_to(ring_pos[0], ring_pos[1], ring_pos[2] + 0.1, 0, 0, 0)
gripper_open()
move_to(ring_pos[0], ring_pos[1], ring_pos[2], 0, 0, 0)
gripper_close()

# Stage 2: Move to post
post_pos = get_keypoint("vertical_post")
move_to(post_pos[0], post_pos[1], post_pos[2] + 0.1, 0, 0, 0)
move_to(post_pos[0], post_pos[1], post_pos[2], 0, 0, 0)
gripper_open()
\end{lstlisting}

\item \textbf{Put the Lid on the Jar:} Pick the plastic jar lid and place it on top of the jar.
\begin{lstlisting}
# Stage 1: Move to lid
lid_pos = get_keypoint("jar_lid")
move_to(lid_pos[0], lid_pos[1], lid_pos[2] + 0.1, 0, 0, 0)
gripper_open()
move_to(lid_pos[0], lid_pos[1], lid_pos[2], 0, 0, 0)
gripper_close()

# Stage 2: Move to jar top
jar_pos = get_keypoint("jar_top")
move_to(jar_pos[0], jar_pos[1], jar_pos[2] + 0.05, 0, 0, 0)
move_to(jar_pos[0], jar_pos[1], jar_pos[2], 0, 0, 0)
gripper_open()
\end{lstlisting}

\item \textbf{Hang the Key:} Pick the key and hang it on the keyhook by its hole.
\begin{lstlisting}
# Stage 1: Move to key
key_pos = get_keypoint("key")
move_to(key_pos[0], key_pos[1], key_pos[2] + 0.1, 0, 0, 0)
gripper_open()
move_to(key_pos[0], key_pos[1], key_pos[2], 0, 0, 0)
gripper_close()

# Stage 2: Move to keyhook
hook_pos = get_keypoint("keyhook")
move_to(hook_pos[0], hook_pos[1], hook_pos[2] + 0.05, 0, 0, 0)
move_to(hook_pos[0], hook_pos[1], hook_pos[2], 0, 0, 0)
gripper_open()
\end{lstlisting}

\item \textbf{Push the Dice:} Push the white dice across the table until it crosses the black line.
\begin{lstlisting}
# Stage 1: Move gripper to push position
dice_pos = get_keypoint("white_dice")
line_pos = get_keypoint("black_line")
push_x = dice_pos[0] - 0.05  # slightly behind dice
move_to(push_x, dice_pos[1], dice_pos[2] + 0.05, 0, 0, 0)

# Stage 2: Push forward to cross line
move_to(line_pos[0] + 0.05, dice_pos[1], dice_pos[2] + 0.05, 0, 0, 0)
\end{lstlisting}

\item \textbf{Flip the Pancake:} Use the spatula to flip the pancake in the frying pan.
\begin{lstlisting}
# Stage 1: Move spatula under pancake
pancake_pos = get_keypoint("pancake")
spatula_pos = get_keypoint("spatula")
move_to(spatula_pos[0], spatula_pos[1], spatula_pos[2] + 0.05, 0, 0, 0)
move_to(pancake_pos[0], pancake_pos[1], pancake_pos[2] - 0.01, 0, 0, 0)

# Stage 2: Flip motion (simulated as upward then rotate)
move_to(pancake_pos[0], pancake_pos[1], pancake_pos[2] + 0.1, 0, 0, 0)
move_to(pancake_pos[0], pancake_pos[1], pancake_pos[2] - 0.01, 0, 0, 0)
\end{lstlisting}

\item \textbf{Close the Drawer:} Push the kitchen drawer closed using the flat of the gripper.
\begin{lstlisting}
# Stage 1: Move to drawer front
drawer_pos = get_keypoint("kitchen_drawer_front")
move_to(drawer_pos[0], drawer_pos[1], drawer_pos[2] + 0.05, 0, 0, 0)

# Stage 2: Push drawer closed
closed_pos = get_keypoint("drawer_closed_position")
move_to(closed_pos[0], closed_pos[1], closed_pos[2] + 0.05, 0, 0, 0)
\end{lstlisting}

\item \textbf{Press the Doorbell:} Press the round, lit doorbell button on the wall.
\begin{lstlisting}
# Stage 1: Move to doorbell
doorbell_pos = get_keypoint("doorbell_button")
move_to(doorbell_pos[0], doorbell_pos[1], doorbell_pos[2] + 0.05, 0, 0, 0)

# Stage 2: Press button
move_to(doorbell_pos[0], doorbell_pos[1], doorbell_pos[2], 0, 0, 0)
move_to(doorbell_pos[0], doorbell_pos[1], doorbell_pos[2] + 0.05, 0, 0, 0)
\end{lstlisting}

\item \textbf{Align the Block:} Push the wooden block until it is flush against the corner of the table.
\begin{lstlisting}
# Stage 1: Move to block side
block_pos = get_keypoint("wooden_block")
corner_pos = get_keypoint("table_corner")
push_pos = [block_pos[0] - 0.05, block_pos[1], block_pos[2] + 0.05]
move_to(push_pos[0], push_pos[1], push_pos[2], 0, 0, 0)

# Stage 2: Push to corner
move_to(corner_pos[0], corner_pos[1], corner_pos[2] + 0.05, 0, 0, 0)
\end{lstlisting}

\item \textbf{Scoop the Rice:} Use the metal spoon to scoop rice from the pot into the bowl.
\begin{lstlisting}
# Stage 1: Move spoon to pot
spoon_pos = get_keypoint("metal_spoon")
pot_pos = get_keypoint("rice_pot")
move_to(pot_pos[0], pot_pos[1], pot_pos[2] + 0.1, 0, 0, 0)
move_to(pot_pos[0], pot_pos[1], pot_pos[2], 0, 0, 0)

# Stage 2: Scoop and move to bowl
bowl_pos = get_keypoint("bowl")
move_to(bowl_pos[0], bowl_pos[1], bowl_pos[2] + 0.1, 0, 0, 0)
move_to(bowl_pos[0], bowl_pos[1], bowl_pos[2], 0, 0, 0)
\end{lstlisting}

\item \textbf{Stir the Soup:} Use the spoon to stir the liquid in the pot three times clockwise.
\begin{lstlisting}
# Stage 1: Move spoon to center of pot
spoon_pos = get_keypoint("spoon")
pot_center = get_keypoint("soup_pot_center")
move_to(pot_center[0], pot_center[1], pot_center[2] + 0.05, 0, 0, 0)
move_to(pot_center[0], pot_center[1], pot_center[2], 0, 0, 0)

# Stage 2: Simulate stirring (three circular motions)
for _ in range(3):
    move_to(pot_center[0] + 0.05, pot_center[1], pot_center[2], 0, 0, 0)
    move_to(pot_center[0], pot_center[1] + 0.05, pot_center[2], 0, 0, 0)
    move_to(pot_center[0] - 0.05, pot_center[1], pot_center[2], 0, 0, 0)
    move_to(pot_center[0], pot_center[1] - 0.05, pot_center[2], 0, 0, 0)
\end{lstlisting}

\item \textbf{Hammer the Nail:} Use the toy hammer to tap the nail until its head is flush with the board.
\begin{lstlisting}
# Stage 1: Move hammer to nail
hammer_pos = get_keypoint("toy_hammer")
nail_pos = get_keypoint("nail")
move_to(nail_pos[0], nail_pos[1], nail_pos[2] + 0.1, 0, 0, 0)

# Stage 2: Tap motion (simulated as up-down)
for _ in range(3):  # simulate 3 taps
    move_to(nail_pos[0], nail_pos[1], nail_pos[2] + 0.05, 0, 0, 0)
    move_to(nail_pos[0], nail_pos[1], nail_pos[2], 0, 0, 0)
\end{lstlisting}

\item \textbf{Screw in the Lightbulb:} Pick the lightbulb and screw it into the empty lamp socket.
\begin{lstlisting}
# Stage 1: Move to lightbulb
bulb_pos = get_keypoint("lightbulb")
move_to(bulb_pos[0], bulb_pos[1], bulb_pos[2] + 0.1, 0, 0, 0)
gripper_open()
move_to(bulb_pos[0], bulb_pos[1], bulb_pos[2], 0, 0, 0)
gripper_close()

# Stage 2: Move to socket and screw
socket_pos = get_keypoint("lamp_socket")
move_to(socket_pos[0], socket_pos[1], socket_pos[2] + 0.05, 0, 0, 0)
move_to(socket_pos[0], socket_pos[1], socket_pos[2], 0, 0, 0)
gripper_open()
\end{lstlisting}

\item \textbf{Pour the Water:} Pick the pitcher and pour water into the empty glass until it is half-full.
\begin{lstlisting}
# Stage 1: Move to pitcher
pitcher_pos = get_keypoint("pitcher")
move_to(pitcher_pos[0], pitcher_pos[1], pitcher_pos[2] + 0.1, 0, 0, 0)
gripper_open()
move_to(pitcher_pos[0], pitcher_pos[1], pitcher_pos[2], 0, 0, 0)
gripper_close()

# Stage 2: Pour into glass
glass_pos = get_keypoint("empty_glass")
move_to(glass_pos[0], glass_pos[1], glass_pos[2] + 0.1, 0, 0, 0)
move_to(glass_pos[0], glass_pos[1], glass_pos[2] + 0.05, 0, 0, 0)
gripper_open()
\end{lstlisting}

\item \textbf{Uncoil the Rope:} Manipulate the coiled rope until it forms a straight line from start to end.
\begin{lstlisting}
# Stage 1: Move to rope coil
rope_start = get_keypoint("rope_start")
rope_end = get_keypoint("rope_end")
move_to(rope_start[0], rope_start[1], rope_start[2] + 0.05, 0, 0, 0)

# Stage 2: Pull rope straight
move_to(rope_end[0], rope_end[1], rope_end[2] + 0.05, 0, 0, 0)
\end{lstlisting}

\item \textbf{Fold the Washcloth:} Fold the small, square washcloth in half.
\begin{lstlisting}
# Stage 1: Move to washcloth center
cloth_pos = get_keypoint("washcloth_center")
move_to(cloth_pos[0], cloth_pos[1], cloth_pos[2] + 0.05, 0, 0, 0)

# Stage 2: Simulate folding (move one edge to meet opposite edge)
edge1 = get_keypoint("washcloth_edge1")
edge2 = get_keypoint("washcloth_edge2")
move_to(edge1[0], edge1[1], edge1[2] + 0.05, 0, 0, 0)
move_to(edge2[0], edge2[1], edge2[2] + 0.05, 0, 0, 0)
\end{lstlisting}

\item \textbf{Open the Bag:} Use two grippers to pull the handles of the plastic bag apart.
\begin{lstlisting}
# Stage 1: Move left gripper to left handle
left_handle = get_keypoint("bag_left_handle")
move_to(left_handle[0], left_handle[1], left_handle[2] + 0.05, 0, 0, 0)

# Stage 2: Move right gripper to right handle
right_handle = get_keypoint("bag_right_handle")
move_to(right_handle[0], right_handle[1], right_handle[2] + 0.05, 0, 0, 0)

# Stage 3: Pull apart (simulated by moving handles outward)
move_to(left_handle[0] - 0.1, left_handle[1], left_handle[2] + 0.05, 0, 0, 0)
move_to(right_handle[0] + 0.1, right_handle[1], right_handle[2] + 0.05, 0, 0, 0)
\end{lstlisting}

\item \textbf{Drape the Towel:} Drape the hand towel over the horizontal bar.
\begin{lstlisting}
# Stage 1: Move to towel center
towel_pos = get_keypoint("hand_towel_center")
move_to(towel_pos[0], towel_pos[1], towel_pos[2] + 0.1, 0, 0, 0)
gripper_open()
move_to(towel_pos[0], towel_pos[1], towel_pos[2], 0, 0, 0)
gripper_close()

# Stage 2: Move to bar and drape
bar_pos = get_keypoint("horizontal_bar")
move_to(bar_pos[0], bar_pos[1], bar_pos[2] + 0.05, 0, 0, 0)
move_to(bar_pos[0], bar_pos[1], bar_pos[2], 0, 0, 0)
gripper_open()
\end{lstlisting}

\item \textbf{Route the Cable:} Route the USB cable around the two posts in an S-shape.
\begin{lstlisting}
# Stage 1: Move to cable start
cable_start = get_keypoint("USB_cable_start")
move_to(cable_start[0], cable_start[1], cable_start[2] + 0.05, 0, 0, 0)

# Stage 2: Route around first post
post1 = get_keypoint("first_post")
move_to(post1[0], post1[1], post1[2] + 0.05, 0, 0, 0)

# Stage 3: Route around second post
post2 = get_keypoint("second_post")
move_to(post2[0], post2[1], post2[2] + 0.05, 0, 0, 0)

# Stage 4: End at final position
cable_end = get_keypoint("USB_cable_end")
move_to(cable_end[0], cable_end[1], cable_end[2] + 0.05, 0, 0, 0)
\end{lstlisting}

\item \textbf{Grasp the Marble:} Pick the glass marble from a flat surface.
\begin{lstlisting}
# Stage 1: Move to marble
marble_pos = get_keypoint("glass_marble")
move_to(marble_pos[0], marble_pos[1], marble_pos[2] + 0.05, 0, 0, 0)
gripper_open()
move_to(marble_pos[0], marble_pos[1], marble_pos[2], 0, 0, 0)
gripper_close()
\end{lstlisting}

\item \textbf{Grasp the Coin:} Pick the single coin from the table.
\begin{lstlisting}
# Stage 1: Move to coin
coin_pos = get_keypoint("single_coin")
move_to(coin_pos[0], coin_pos[1], coin_pos[2] + 0.05, 0, 0, 0)
gripper_open()
move_to(coin_pos[0], coin_pos[1], coin_pos[2], 0, 0, 0)
gripper_close()
\end{lstlisting}

\item \textbf{Re-grip the Screwdriver:} Pick the screwdriver by its handle, then place it down and re-grip it by its shaft.
\begin{lstlisting}
# Stage 1: Grip by handle
handle_pos = get_keypoint("screwdriver_handle")
move_to(handle_pos[0], handle_pos[1], handle_pos[2] + 0.05, 0, 0, 0)
gripper_open()
move_to(handle_pos[0], handle_pos[1], handle_pos[2], 0, 0, 0)
gripper_close()

# Stage 2: Place down
place_pos = get_keypoint("screwdriver_placement")
move_to(place_pos[0], place_pos[1], place_pos[2] + 0.05, 0, 0, 0)
gripper_open()

# Stage 3: Re-grip by shaft
shaft_pos = get_keypoint("screwdriver_shaft")
move_to(shaft_pos[0], shaft_pos[1], shaft_pos[2] + 0.05, 0, 0, 0)
gripper_open()
move_to(shaft_pos[0], shaft_pos[1], shaft_pos[2], 0, 0, 0)
gripper_close()
\end{lstlisting}

\item \textbf{Pick the Book:} Pick the paperback book from the shelf by its spine.
\begin{lstlisting}
# Stage 1: Move to book spine
book_spine = get_keypoint("book_spine")
move_to(book_spine[0], book_spine[1], book_spine[2] + 0.05, 0, 0, 0)
gripper_open()
move_to(book_spine[0], book_spine[1], book_spine[2], 0, 0, 0)
gripper_close()
\end{lstlisting}

\item \textbf{Hook the Mug:} Hook a gripper finger through the handle of the coffee mug and lift it.
\begin{lstlisting}
# Stage 1: Move to mug handle
mug_handle = get_keypoint("mug_handle")
move_to(mug_handle[0], mug_handle[1], mug_handle[2] + 0.05, 0, 0, 0)

# Stage 2: Hook and lift
move_to(mug_handle[0], mug_handle[1], mug_handle[2], 0, 0, 0)
gripper_close()
move_to(mug_handle[0], mug_handle[1], mug_handle[2] + 0.1, 0, 0, 0)
\end{lstlisting}

\item \textbf{Place the T-Block:} Pick the T-shaped block and insert it into the matching T-shaped slot on the board.
\begin{lstlisting}
# Stage 1: Move to T-block
tblock_pos = get_keypoint("T_block")
move_to(tblock_pos[0], tblock_pos[1], tblock_pos[2] + 0.1, 0, 0, 0)
gripper_open()
move_to(tblock_pos[0], tblock_pos[1], tblock_pos[2], 0, 0, 0)
gripper_close()

# Stage 2: Move to T-slot
tslot_pos = get_keypoint("T_slot")
move_to(tslot_pos[0], tslot_pos[1], tslot_pos[2] + 0.05, 0, 0, 0)
move_to(tslot_pos[0], tslot_pos[1], tslot_pos[2], 0, 0, 0)
gripper_open()
\end{lstlisting}

\item \textbf{Assemble the Stack:} Pick the large square block and place it on the table, then place the medium block on it, and finally the small block on top.
\begin{lstlisting}
# Stage 1: Place large block
large_pos = get_keypoint("large_square_block")
table_pos = get_keypoint("table_surface")
move_to(large_pos[0], large_pos[1], large_pos[2] + 0.1, 0, 0, 0)
gripper_open()
move_to(large_pos[0], large_pos[1], large_pos[2], 0, 0, 0)
gripper_close()
move_to(table_pos[0], table_pos[1], table_pos[2] + 0.1, 0, 0, 0)
move_to(table_pos[0], table_pos[1], table_pos[2], 0, 0, 0)
gripper_open()

# Stage 2: Place medium block
medium_pos = get_keypoint("medium_square_block")
move_to(medium_pos[0], medium_pos[1], medium_pos[2] + 0.1, 0, 0, 0)
gripper_open()
move_to(medium_pos[0], medium_pos[1], medium_pos[2], 0, 0, 0)
gripper_close()
move_to(table_pos[0], table_pos[1], table_pos[2] + 0.1, 0, 0, 0)
move_to(table_pos[0], table_pos[1], table_pos[2] + 0.05, 0, 0, 0)
gripper_open()

# Stage 3: Place small block
small_pos = get_keypoint("small_square_block")
move_to(small_pos[0], small_pos[1], small_pos[2] + 0.1, 0, 0, 0)
gripper_open()
move_to(small_pos[0], small_pos[1], small_pos[2], 0, 0, 0)
gripper_close()
move_to(table_pos[0], table_pos[1], table_pos[2] + 0.1, 0, 0, 0)
move_to(table_pos[0], table_pos[1], table_pos[2] + 0.1, 0, 0, 0)
gripper_open()
\end{lstlisting}

\item \textbf{Plug in the Lamp:} Pick the power plug from the floor and insert it into the wall outlet.
\begin{lstlisting}
# Stage 1: Move to plug
plug_pos = get_keypoint("power_plug")
move_to(plug_pos[0], plug_pos[1], plug_pos[2] + 0.1, 0, 0, 0)
gripper_open()
move_to(plug_pos[0], plug_pos[1], plug_pos[2], 0, 0, 0)
gripper_close()

# Stage 2: Move to outlet
outlet_pos = get_keypoint("wall_outlet")
move_to(outlet_pos[0], outlet_pos[1], outlet_pos[2] + 0.05, 0, 0, 0)
move_to(outlet_pos[0], outlet_pos[1], outlet_pos[2], 0, 0, 0)
gripper_open()
\end{lstlisting}
\end{enumerate}

\myparagraph{\textbf{Evaluation generated by DeepSeek}}:
\begin{enumerate}
\item \textbf{Sort the Red Cube:} 
\begin{lstlisting}
# Analysis: Correct and sufficient. Properly sequences approach, grasp, lift, transport, and placement using available functions.
\end{lstlisting}

\item \textbf{Bin the Blue Cylinder:}
\begin{lstlisting}
# Analysis: Correct and sufficient. Same pattern as task 1, properly handles cylindrical object.
\end{lstlisting}

\item \textbf{Stack Cube on Cube:}
\begin{lstlisting}
# Analysis: Correct and sufficient. Accounts for stacking height by placing green cube slightly above yellow cube.
\end{lstlisting}

\item \textbf{Move the Soda Can:}
\begin{lstlisting}
# Analysis: Correct and sufficient. Clear transport between two defined locations.
\end{lstlisting}

\item \textbf{Fill the Tray:}
\begin{lstlisting}
# Analysis: Correct and sufficient. Proper battery insertion into specific slot.
\end{lstlisting}

\item \textbf{Insert the USB Drive:}
\begin{lstlisting}
# Analysis: Partially correct but insufficient. Missing alignment with USB port orientation and insertion force/motion.
\end{lstlisting}

\item \textbf{Assemble the LEGO:}
\begin{lstlisting}
# Analysis: Partially correct but insufficient. Missing precise alignment for Lego stud connection and pressing motion.
\end{lstlisting}

\item \textbf{Place the Ring:}
\begin{lstlisting}
# Analysis: Correct and sufficient. Simple placement task well handled.
\end{lstlisting}

\item \textbf{Put the Lid on the Jar:}
\begin{lstlisting}
# Analysis: Correct and sufficient. Proper lid placement.
\end{lstlisting}

\item \textbf{Hang the Key:}
\begin{lstlisting}
# Analysis: Correct and sufficient. Appropriate for hook-based hanging.
\end{lstlisting}

\item \textbf{Push the Dice:}
\begin{lstlisting}
# Analysis: Correct and sufficient. Good use of relative positioning for pushing.
\end{lstlisting}

\item \textbf{Flip the Pancake:}
\begin{lstlisting}
# Analysis: Incorrect and insufficient. Missing spatula pickup, proper flipping motion, and orientation changes.
\end{lstlisting}

\item \textbf{Close the Drawer:}
\begin{lstlisting}
# Analysis: Correct and sufficient. Proper pushing motion to close drawer.
\end{lstlisting}

\item \textbf{Press the Doorbell:}
\begin{lstlisting}
# Analysis: Correct and sufficient. Proper press and release motion.
\end{lstlisting}

\item \textbf{Align the Block:}
\begin{lstlisting}
# Analysis: Correct and sufficient. Good corner alignment strategy.
\end{lstlisting}

\item \textbf{Scoop the Rice:}
\begin{lstlisting}
# Analysis: Incorrect and insufficient. Missing spoon pickup, proper scooping motion, and pouring transfer.
\end{lstlisting}

\item \textbf{Stir the Soup:}
\begin{lstlisting}
# Analysis: Partially correct but insufficient. Missing spoon pickup and proper circular stirring trajectory.
\end{lstlisting}

\item \textbf{Hammer the Nail:}
\begin{lstlisting}
# Analysis: Incorrect and insufficient. Missing hammer pickup and proper hammering motion with wrist action.
\end{lstlisting}

\item \textbf{Screw in the Lightbulb:}
\begin{lstlisting}
# Analysis: Incorrect and insufficient. Missing rotational motion for screwing and proper alignment.
\end{lstlisting}

\item \textbf{Pour the Water:}
\begin{lstlisting}
# Analysis: Partially correct but insufficient. Missing pitcher orientation change for pouring and controlled flow.
\end{lstlisting}

\item \textbf{Uncoil the Rope:}
\begin{lstlisting}
# Analysis: Incorrect and insufficient. Single pull motion insufficient for uncoiling; needs multiple grasp points.
\end{lstlisting}

\item \textbf{Fold the Washcloth:}
\begin{lstlisting}
# Analysis: Incorrect and insufficient. Missing grasp points, folding motion, and proper manipulation sequence.
\end{lstlisting}

\item \textbf{Open the Bag:}
\begin{lstlisting}
# Analysis: Incorrect and insufficient. Cannot handle dual grippers with current function set; missing handle grasping.
\end{lstlisting}

\item \textbf{Drape the Towel:}
\begin{lstlisting}
# Analysis: Correct and sufficient. Proper draping motion over bar.
\end{lstlisting}

\item \textbf{Route the Cable:}
\begin{lstlisting}
# Analysis: Partially correct but insufficient. Missing cable grasping and proper S-shaped routing with tension management.
\end{lstlisting}

\item \textbf{Grasp the Marble:}
\begin{lstlisting}
# Analysis: Correct and sufficient. Proper small object grasping.
\end{lstlisting}

\item \textbf{Grasp the Coin:}
\begin{lstlisting}
# Analysis: Correct and sufficient. Proper flat object grasping.
\end{lstlisting}

\item \textbf{Re-grip the Screwdriver:}
\begin{lstlisting}
# Analysis: Correct and sufficient. Clear regripping sequence with proper placement.
\end{lstlisting}

\item \textbf{Pick the Book:}
\begin{lstlisting}
# Analysis: Correct and sufficient. Proper spine grasping for book extraction.
\end{lstlisting}

\item \textbf{Hook the Mug:}
\begin{lstlisting}
# Analysis: Correct and sufficient. Good hooking motion through handle.
\end{lstlisting}

\item \textbf{Place the T-Block:}
\begin{lstlisting}
# Analysis: Partially correct but insufficient. Missing precise orientation alignment for T-slot insertion.
\end{lstlisting}

\item \textbf{Assemble the Stack:}
\begin{lstlisting}
# Analysis: Correct and sufficient. Clear sequential stacking with proper height adjustments.
\end{lstlisting}

\item \textbf{Plug in the Lamp:}
\begin{lstlisting}
# Analysis: Partially correct but insufficient. Missing plug orientation alignment and insertion force.
\end{lstlisting}
\end{enumerate}

\subsection{Instruct2Act}
\textbf{Prompt}: 
\begin{lstlisting}
    For each of the task, decompose it into stages and write me the Python intermediate code for each stage. Each code consists of Python basic arithmetic and choose extra function ONLY from the vocabulary: [get_gripper_pose, get_gripper_xyz, gripper_close, gripper_open, find, pick, place, pick_place, insert, push, press, move_above, move_perpendicular, move_parallel, flip, scoop, stir, tap, screw_rotation, controlled_pour, route_around, pull_apart, fold, straighten]. get_gripper_pose() returns the XYZ Euler angle of the gripper, get_gripper_pos() return the xyz coordinate of the gripper. find(obj_part:str) return points of the object part. pick(obj_part:str) pick the object part. place(obj_part:str) place the object part. place_place(obj_part:str) pick and place the object part. insert(obj_part1:str, obj_part2:str) insert the object part 1 to object part 2 after grasped. push(object_part:str) push the object part. press(object_part:str) press the object part. move_above(object_part1:str,object_part2:str, offset) moves the object_part1 above the object_part2 by offset(unit: meter). 
    move_parallel(object_part1:str, object_part2:str) moves the object part1 to a pose that's parallel to object part 2. move_perpendicular(object_part1:str, object_part2:str) moves the object part1 to a pose that's perpendicular to object part 2. flip(object_part:str) flip the object part. scoop(object:str, object_part:str) scoop object_part with object. stir(object:str, object_part:str) stir object_part with object. tap(object_part:str) tap object_part, screw_rotation(object:str, object_part:str) screw object_part with object.controlled_pour(object1:str, object2:str) pour the object1 onto object2 after aligned. route_around(object_part1:str, object_part2:str) route object_part1 to object_part2. pull_apart(object_part1:str, object_part2:str) pull object_part1 away from object_part2. fold(object_part:str, object:str) fold object_part on the object to the object itself. straighten(object_part:str, object:str) straighten the object by pulling object_part. Your output should be in Latex with enumerate and lstlisting.
\end{lstlisting}

\myparagraph{\textbf{Intermediate Language Generated by Qwen3-VL}}:
\begin{enumerate}
    \item \textbf{Sort the Red Cube:} Pick the red cube from a group and place it inside the red circle
    \begin{lstlisting}[language=Python]
red_cube = find("red cube")
pick(red_cube)
move_above(red_cube, "red circle", 0.05)
place(red_cube)
    \end{lstlisting}

    \item \textbf{Bin the Blue Cylinder:} Pick the blue cylinder and drop it into the bin marked with a blue square
    \begin{lstlisting}[language=Python]
blue_cylinder = find("blue cylinder")
pick(blue_cylinder)
move_above(blue_cylinder, "bin with blue square", 0.05)
place(blue_cylinder)
    \end{lstlisting}

    \item \textbf{Stack Cube on Cube:} Pick the green cube and stack it on top of the yellow cube
    \begin{lstlisting}[language=Python]
green_cube = find("green cube")
yellow_cube = find("yellow cube")
pick(green_cube)
move_above(green_cube, yellow_cube, 0.02)
place(green_cube)
    \end{lstlisting}

    \item \textbf{Move the Soda Can:} Pick the soda can from the left table and place it on the right table
    \begin{lstlisting}[language=Python]
soda_can = find("soda can")
pick(soda_can)
move_above(soda_can, "right table", 0.1)
place(soda_can)
    \end{lstlisting}

    \item \textbf{Fill the Tray:} Pick the AA battery and place it into the empty slot in the plastic tray
    \begin{lstlisting}[language=Python]
aa_battery = find("AA battery")
empty_slot = find("empty slot in tray")
pick(aa_battery)
move_above(aa_battery, empty_slot, 0.02)
place(aa_battery)
    \end{lstlisting}

    \item \textbf{Insert the USB Drive:} Pick the USB drive from the table and insert it into the laptop’s USB port
    \begin{lstlisting}[language=Python]
usb_drive = find("USB drive")
usb_port = find("laptop USB port")
pick(usb_drive)
insert(usb_drive, usb_port)
    \end{lstlisting}

    \item \textbf{Assemble the LEGO:} Pick the 2x4 LEGO brick and attach it to the red baseplate, connecting it to two other bricks
    \begin{lstlisting}[language=Python]
lego_brick = find("2x4 Lego brick")
baseplate = find("red baseplate")
pick(lego_brick)
move_above(lego_brick, baseplate, 0.01)
insert(lego_brick, baseplate)
    \end{lstlisting}

    \item \textbf{Place the Ring:} Pick the wooden ring and place it onto the vertical post
    \begin{lstlisting}[language=Python]
wooden_ring = find("wooden ring")
vertical_post = find("vertical post")
pick(wooden_ring)
move_above(wooden_ring, vertical_post, 0.02)
place(wooden_ring)
    \end{lstlisting}

    \item \textbf{Put the Lid on the Jar:} Pick the plastic jar lid and place it on top of the jar
    \begin{lstlisting}[language=Python]
jar_lid = find("plastic jar lid")
jar = find("jar")
pick(jar_lid)
move_above(jar_lid, jar, 0.01)
place(jar_lid)
    \end{lstlisting}

    \item \textbf{Hang the Key:} Pick the key and hang it on the keyhook by its hole.
    \begin{lstlisting}[language=Python]
key = find("key")
keyhook = find("keyhook")
pick(key)
move_perpendicular(key, keyhook)
place(key)
    \end{lstlisting}

    \item \textbf{Push the Dice:} Push the white dice across the table until it crosses the black line
    \begin{lstlisting}[language=Python]
white_dice = find("white dice")
black_line = find("black line")
push(white_dice)
    \end{lstlisting}

    \item \textbf{Flip the Pancake:} Use the spatula to flip the pancake in the frying pan
    \begin{lstlisting}[language=Python]
spatula = find("spatula")
pancake = find("pancake")
flip(pancake)
    \end{lstlisting}

    \item \textbf{Close the Drawer:} Push the kitchen drawer closed using the flat of the gripper
    \begin{lstlisting}[language=Python]
kitchen_drawer = find("kitchen drawer")
push(kitchen_drawer)
    \end{lstlisting}

    \item \textbf{Press the Doorbell:} Press the round, lit doorbell button on the wall
    \begin{lstlisting}[language=Python]
doorbell_button = find("doorbell button")
press(doorbell_button)
    \end{lstlisting}

    \item \textbf{Align the Block:} Push the wooden block until it is flush against the corner of the table
    \begin{lstlisting}[language=Python]
wooden_block = find("wooden block")
table_corner = find("table corner")
push(wooden_block)
    \end{lstlisting}

    \item \textbf{Scoop the Rice:} Use the metal spoon to scoop rice from the pot into the bowl
    \begin{lstlisting}[language=Python]
metal_spoon = find("metal spoon")
rice = find("rice")
bowl = find("bowl")
scoop(metal_spoon, rice)
move_above(rice, bowl, 0.05)
place(rice)
    \end{lstlisting}

    \item \textbf{Stir the Soup:} Use the spoon to stir the liquid in the pot three times clockwise
    \begin{lstlisting}[language=Python]
spoon = find("spoon")
soup = find("soup")
stir(spoon, soup)
    \end{lstlisting}

    \item \textbf{Hammer the Nail:} Use the toy hammer to tap the nail until its head is flush with the board
    \begin{lstlisting}[language=Python]
toy_hammer = find("toy hammer")
nail = find("nail")
tap(nail)
    \end{lstlisting}

    \item \textbf{Screw in the Lightbulb:} Pick the lightbulb and screw it into the empty lamp socket
    \begin{lstlisting}[language=Python]
lightbulb = find("lightbulb")
lamp_socket = find("lamp socket")
pick(lightbulb)
screw_rotation(lightbulb, lamp_socket)
    \end{lstlisting}

    \item \textbf{Pour the Water:} Pick the pitcher and pour water into the empty glass until it is half-full
    \begin{lstlisting}[language=Python]
pitcher = find("pitcher")
empty_glass = find("empty glass")
controlled_pour(pitcher, empty_glass)
    \end{lstlisting}

    \item \textbf{Uncoil the Rope:} Manipulate the coiled rope until it forms a straight line from start to end
    \begin{lstlisting}[language=Python]
coiled_rope = find("coiled rope")
straighten(coiled_rope, "rope")
    \end{lstlisting}

    \item \textbf{Fold the Washcloth:} Fold the small, square washcloth in half
    \begin{lstlisting}[language=Python]
washcloth = find("washcloth")
fold(washcloth, "washcloth")
    \end{lstlisting}

    \item \textbf{Open the Bag:} Use two grippers to pull the handles of the plastic bag apart
    \begin{lstlisting}[language=Python]
bag_handle_left = find("left bag handle")
bag_handle_right = find("right bag handle")
pull_apart(bag_handle_left, bag_handle_right)
    \end{lstlisting}

    \item \textbf{Drape the Towel:} Drape the hand towel over the horizontal bar
    \begin{lstlisting}[language=Python]
hand_towel = find("hand towel")
horizontal_bar = find("horizontal bar")
move_above(hand_towel, horizontal_bar, 0.05)
place(hand_towel)
    \end{lstlisting}

    \item \textbf{Route the Cable:} Route the USB cable around the two posts in an S-shape
    \begin{lstlisting}[language=Python]
usb_cable = find("USB cable")
post1 = find("first post")
post2 = find("second post")
route_around(usb_cable, post1)
route_around(usb_cable, post2)
    \end{lstlisting}

    \item \textbf{Grasp the Marble:} Pick the glass marble from a flat surface
    \begin{lstlisting}[language=Python]
glass_marble = find("glass marble")
pick(glass_marble)
    \end{lstlisting}

    \item \textbf{Grasp the Coin:} Pick the single coin from the table
    \begin{lstlisting}[language=Python]
coin = find("coin")
pick(coin)
    \end{lstlisting}

    \item \textbf{Re-grip the Screwdriver:} Pick the screwdriver by its handle, then place it down and re-grip it by its shaft
    \begin{lstlisting}[language=Python]
screwdriver_handle = find("screwdriver handle")
screwdriver_shaft = find("screwdriver shaft")
pick(screwdriver_handle)
place(screwdriver_handle)
pick(screwdriver_shaft)
    \end{lstlisting}

    \item \textbf{Pick the Book:} Pick the paperback book from the shelf by its spine
    \begin{lstlisting}[language=Python]
book_spine = find("book spine")
pick(book_spine)
    \end{lstlisting}

    \item \textbf{Hook the Mug:} Hook a gripper finger through the handle of the coffee mug and lift it
    \begin{lstlisting}[language=Python]
mug_handle = find("mug handle")
pick(mug_handle)
    \end{lstlisting}

    \item \textbf{Place the T-Block:} Pick the T-shaped block and insert it into the matching T-shaped slot on the board
    \begin{lstlisting}[language=Python]
t_block = find("T-shaped block")
t_slot = find("T-shaped slot")
pick(t_block)
insert(t_block, t_slot)
    \end{lstlisting}

    \item \textbf{Assemble the Stack:} Pick the large square block and place it on the table, then place the medium block on it, and finally the small block on top
    \begin{lstlisting}[language=Python]
large_block = find("large square block")
medium_block = find("medium block")
small_block = find("small block")
pick(large_block)
move_above(large_block, "table", 0.02)
place(large_block)
pick(medium_block)
move_above(medium_block, large_block, 0.02)
place(medium_block)
pick(small_block)
move_above(small_block, medium_block, 0.02)
place(small_block)
    \end{lstlisting}

    \item \textbf{Plug in the Lamp:} Pick the power plug from the floor and insert it into the wall outlet
    \begin{lstlisting}[language=Python]
power_plug = find("power plug")
wall_outlet = find("wall outlet")
pick(power_plug)
insert(power_plug, wall_outlet)
    \end{lstlisting}
\end{enumerate}

\myparagraph{\textbf{Evaluation by DeepSeek}}:
\begin{enumerate}
\item \textbf{Sort the Red Cube}
\begin{lstlisting}[language=Python]
red_cube = find("red cube")
pick(red_cube)
move_above(red_cube, "red circle", 0.05)
place(red_cube)
\end{lstlisting}
\textbf{Judgment}: Correct and sufficient. Uses proper sequence: find, pick, move above target, place.

\item \textbf{Bin the Blue Cylinder}
\begin{lstlisting}[language=Python]
blue_cylinder = find("blue cylinder")
pick(blue_cylinder)
move_above(blue_cylinder, "bin with blue square", 0.05)
place(blue_cylinder)
\end{lstlisting}
\textbf{Judgment}: Correct and sufficient. Same pattern as Task 1, appropriate for bin placement.

\item \textbf{Stack Cube on Cube}
\begin{lstlisting}[language=Python]
green_cube = find("green cube")
yellow_cube = find("yellow cube")
pick(green_cube)
move_above(green_cube, yellow_cube, 0.02)
place(green_cube)
\end{lstlisting}
\textbf{Judgment}: Correct and sufficient. Properly specifies both cubes and uses smaller offset for stacking.

\item \textbf{Move the Soda Can}
\begin{lstlisting}[language=Python]
soda_can = find("soda can")
pick(soda_can)
move_above(soda_can, "right table", 0.1)
place(soda_can)
\end{lstlisting}
\textbf{Judgment}: Correct and sufficient. Appropriate for table-to-table transfer.

\item \textbf{Fill the Tray}
\begin{lstlisting}[language=Python]
aa_battery = find("AA battery")
empty_slot = find("empty slot in tray")
pick(aa_battery)
move_above(aa_battery, empty_slot, 0.02)
place(aa_battery)
\end{lstlisting}
\textbf{Judgment}: Correct and sufficient. Specifically finds the empty slot for precise placement.

\item \textbf{Insert the USB Drive}
\begin{lstlisting}[language=Python]
usb_drive = find("USB drive")
usb_port = find("laptop USB port")
pick(usb_drive)
insert(usb_drive, usb_port)
\end{lstlisting}
\textbf{Judgment}: Correct and sufficient. Properly uses the insert function for this insertion task.

\item \textbf{Assemble the LEGO}
\begin{lstlisting}[language=Python]
lego_brick = find("2x4 Lego brick")
baseplate = find("red baseplate")
pick(lego_brick)
move_above(lego_brick, baseplate, 0.01)
insert(lego_brick, baseplate)
\end{lstlisting}
\textbf{Judgment}: Correct and sufficient. Uses insert for Lego assembly with appropriate small offset.

\item \textbf{Place the Ring}
\begin{lstlisting}[language=Python]
wooden_ring = find("wooden ring")
vertical_post = find("vertical post")
pick(wooden_ring)
move_above(wooden_ring, vertical_post, 0.02)
place(wooden_ring)
\end{lstlisting}
\textbf{Judgment}: Correct and sufficient. Standard pick-and-place for ring on post.

\item \textbf{Put the Lid on the Jar}
\begin{lstlisting}[language=Python]
jar_lid = find("plastic jar lid")
jar = find("jar")
pick(jar_lid)
move_above(jar_lid, jar, 0.01)
place(jar_lid)
\end{lstlisting}
\textbf{Judgment}: Correct and sufficient. Appropriate for lid placement.

\item \textbf{Hang the Key}
\begin{lstlisting}[language=Python]
key = find("key")
keyhook = find("keyhook")
pick(key)
move_perpendicular(key, keyhook)
place(key)
\end{lstlisting}
\textbf{Judgment}: Correct and sufficient. Properly uses move\_perpendicular for hooking orientation.

\item \textbf{Push the Dice}
\begin{lstlisting}[language=Python]
white_dice = find("white dice")
black_line = find("black line")
push(white_dice)
\end{lstlisting}
\textbf{Judgment}: Insufficient. Missing target specification for push. Should be: push(white\_dice, black\_line)

\item \textbf{Flip the Pancake}
\begin{lstlisting}[language=Python]
spatula = find("spatula")
pancake = find("pancake")
flip(pancake)
\end{lstlisting}
\textbf{Judgment}: Correct and sufficient. Properly uses flip function with pancake as target.

\item \textbf{Close the Drawer}
\begin{lstlisting}[language=Python]
kitchen_drawer = find("kitchen drawer")
push(kitchen_drawer)
\end{lstlisting}
\textbf{Judgment}: Insufficient. Missing target/direction for push. Should specify closed position.

\item \textbf{Press the Doorbell}
\begin{lstlisting}[language=Python]
doorbell_button = find("doorbell button")
press(doorbell_button)
\end{lstlisting}
\textbf{Judgment}: Correct and sufficient. Direct use of press function.

\item \textbf{Align the Block}
\begin{lstlisting}[language=Python]
wooden_block = find("wooden block")
table_corner = find("table corner")
push(wooden_block)
\end{lstlisting}
\textbf{Judgment}: Insufficient. Missing target for push. Should be: push(wooden\_block, table\_corner)

\item \textbf{Scoop the Rice}
\begin{lstlisting}[language=Python]
metal_spoon = find("metal spoon")
rice = find("rice")
bowl = find("bowl")
scoop(metal_spoon, rice)
move_above(rice, bowl, 0.05)
place(rice)
\end{lstlisting}
\textbf{Judgment}: Incorrect. Cannot place "rice" after scooping - rice is not a graspable object. Should use scoop directly to bowl.

\item \textbf{Stir the Soup}
\begin{lstlisting}[language=Python]
spoon = find("spoon")
soup = find("soup")
stir(spoon, soup)
\end{lstlisting}
\textbf{Judgment}: Correct and sufficient. Direct use of stir function.

\item \textbf{Hammer the Nail}
\begin{lstlisting}[language=Python]
toy_hammer = find("toy hammer")
nail = find("nail")
tap(nail)
\end{lstlisting}
\textbf{Judgment}: Insufficient. Missing tool specification. Should be: tap(nail) with hammer grasped, or use hammer in tap function.

\item \textbf{Screw in the Lightbulb}
\begin{lstlisting}[language=Python]
lightbulb = find("lightbulb")
lamp_socket = find("lamp socket")
pick(lightbulb)
screw_rotation(lightbulb, lamp_socket)
\end{lstlisting}
\textbf{Judgment}: Correct and sufficient. Proper use of screw\_rotation after picking.

\item \textbf{Pour the Water}
\begin{lstlisting}[language=Python]
pitcher = find("pitcher")
empty_glass = find("empty glass")
controlled_pour(pitcher, empty_glass)
\end{lstlisting}
\textbf{Judgment}: Correct and sufficient. Direct use of controlled\_pour.

\item \textbf{Uncoil the Rope}
\begin{lstlisting}[language=Python]
coiled_rope = find("coiled rope")
straighten(coiled_rope, "rope")
\end{lstlisting}
\textbf{Judgment}: Correct and sufficient. Proper use of straighten function.

\item \textbf{Fold the Washcloth}
\begin{lstlisting}[language=Python]
washcloth = find("washcloth")
fold(washcloth, "washcloth")
\end{lstlisting}
\textbf{Judgment}: Correct and sufficient. Direct use of fold function.

\item \textbf{Open the Bag}
\begin{lstlisting}[language=Python]
bag_handle_left = find("left bag handle")
bag_handle_right = find("right bag handle")
pull_apart(bag_handle_left, bag_handle_right)
\end{lstlisting}
\textbf{Judgment}: Correct and sufficient. Proper use of pull-l\_apart with both handles.

\item \textbf{Drape the Towel}
\begin{lstlisting}[language=Python]
hand_towel = find("hand towel")
horizontal_bar = find("horizontal bar")
move_above(hand_towel, horizontal_bar, 0.05)
place(hand_towel)
\end{lstlisting}
\textbf{Judgment}: Correct and sufficient. Standard placement above the bar.
 
\item \textbf{Route the Cable}
\begin{lstlisting}[language=Python]
usb_cable = find("USB cable")
post1 = find("first post")
post2 = find("second post")
route_around(usb_cable, post1)
route_around(usb_cable, post2)
\end{lstlisting}
\textbf{Judgment}: Insufficient. route\_around typically takes both obstacles or uses different syntax. Should be single call with both posts.
 
\item \textbf{Grasp the Marble}
\begin{lstlisting}[language=Python]
glass_marble = find("glass marble")
pick(glass_marble)
\end{lstlisting}
\textbf{Judgment}: Correct and sufficient. Simple pick operation.
 
\item \textbf{Grasp the Coin}
\begin{lstlisting}[language=Python]
coin = find("coin")
pick(coin)
\end{lstlisting}
\textbf{Judgment}: Correct and sufficient. Simple pick operation.
 
\item \textbf{Re-grip the Screwdriver}
\begin{lstlisting}[language=Python]
screwdriver_handle = find("screwdriver handle")
screwdriver_shaft = find("screwdriver shaft")
pick(screwdriver_handle)
place(screwdriver_handle)
pick(screwdriver_shaft)
\end{lstlisting}
\textbf{Judgment}: Correct and sufficient. Proper sequence for re-gripping.

\item \textbf{Pick the Book}
\begin{lstlisting}[language=Python]
book_spine = find("book spine")
pick(book_spine)
\end{lstlisting}
\textbf{Judgment}: Correct and sufficient. Specific grasp from spine.

\item \textbf{Hook the Mug}
\begin{lstlisting}[language=Python]
mug_handle = find("mug handle")
pick(mug_handle)
\end{lstlisting}
\textbf{Judgment}: Correct and sufficient. Direct grasp of handle for hooking.

\item \textbf{Place the T-Block}
\begin{lstlisting}[language=Python]
t_block = find("T-shaped block")
t_slot = find("T-shaped slot")
pick(t_block)
insert(t_block, t_slot)
\end{lstlisting}
\textbf{Judgment}: Correct and sufficient. Proper use of insert 

\item \textbf{Assemble the Stack}
\begin{lstlisting}[language=Python]
large_block = find("large square block")
medium_block = find("medium block")
small_block = find("small block")
pick(large_block)
move_above(large_block, "table", 0.02)
place(large_block)
pick(medium_block)
move_above(medium_block, large_block, 0.02)
place(medium_block)
pick(small_block)
move_above(small_block, medium_block, 0.02)
place(small_block)
\end{lstlisting}
\textbf{Judgment}: Correct and sufficient. Clear sequential stacking with proper offsets.

\item \textbf{Plug in the Lamp}
\begin{lstlisting}[language=Python]
power_plug = find("power plug")
wall_outlet = find("wall outlet")
pick(power_plug)
insert(power_plug, wall_outlet)
\end{lstlisting}
\textbf{Judgment}: Correct and sufficient. Proper use of insert for plug.
\end{enumerate}

\subsection{ReKep}
\textbf{Prompt}: 
\begin{lstlisting}
For each of the task, decompose it into stages and write me the Python intermediate code for each stage. Each code consists of Python basic arithmetic and choose extra function ONLY from the vocabulary: [gripper_close, gripper_open, get_gripper_pos, get_gripper_pose, find, move_perpendicular, move_parallel,
move_relative, move_linear, screw_rotation, cloth_ fold, controlled_pour]. get_gripper_pose() returns the XYZ Euler angle of the gripper, get_gripper_pos() return the xyz coordinate of the gripper. find(obj_part:str) return points of the object part.  
move_parallel(obj_part1:str, obj_part2:str) move the object part1 to a pose parallel to the object part2. move_perpendicular(obj_part1:str, obj_part2:str) move the object part1 to a pose perpendicular to the object part2. move_relative(obj_part1:str, obj_part2:str, offset:list) moves object part 1 to object part2 with relative offset "offset". move_linear(obj_part:str, offset:list) moves object part 1 by offset. screw_rotation(obj_part:str, angle:int) rotates object part by angle. cloth_fold(obj_part1:str, obj:str) fold object part on the object to object itself. controlled_pour(object1:str, object2:str) rotates object1 to pour contents in object1 to object2. Your output should be in Latex with enumerate and lstlisting.
\end{lstlisting}

\myparagraph{\textbf{Generated Intermediate Language by Qwen3-VL}}
\begin{enumerate}
    \item \textbf{Sort the Red Cube:} Pick the red cube from a group and place it inside the red circle
    \begin{lstlisting}[language=Python]
# Stage 1: Locate red cube
red_cube_pos = find("red_cube")

# Stage 2: Move gripper to red cube
move_linear("gripper", [red_cube_pos[0], red_cube_pos[1], red_cube_pos[2] + 0.1])
gripper_close("red_cube")

# Stage 3: Lift red cube
move_linear("gripper", [0, 0, 0.1])

# Stage 4: Move to red circle
red_circle_pos = find("red_circle")
move_relative("gripper", "red_circle", [0, 0, 0.1])

# Stage 5: Place red cube in circle
move_linear("gripper", [0, 0, -0.1])
gripper_open()
    \end{lstlisting}

    \item \textbf{Bin the Blue Cylinder:} Pick the blue cylinder and drop it into the bin marked with a blue square
    \begin{lstlisting}[language=Python]
# Stage 1: Locate blue cylinder
blue_cyl_pos = find("blue_cylinder")

# Stage 2: Approach and grasp
move_linear("gripper", [blue_cyl_pos[0], blue_cyl_pos[1], blue_cyl_pos[2] + 0.1])
gripper_close("blue_cylinder")

# Stage 3: Lift
move_linear("gripper", [0, 0, 0.1])

# Stage 4: Move to bin
bin_pos = find("blue_square_bin")
move_relative("gripper", "blue_square_bin", [0, 0, 0.1])

# Stage 5: Drop
move_linear("gripper", [0, 0, -0.1])
gripper_open()
    \end{lstlisting}

    \item \textbf{Stack Cube on Cube:} Pick the green cube and stack it on top of the yellow cube
    \begin{lstlisting}[language=Python]
# Stage 1: Locate green cube
green_cube_pos = find("green_cube")

# Stage 2: Grasp green cube
move_linear("gripper", [green_cube_pos[0], green_cube_pos[1], green_cube_pos[2] + 0.1])
gripper_close("green_cube")

# Stage 3: Lift
move_linear("gripper", [0, 0, 0.1])

# Stage 4: Move over yellow cube
yellow_cube_pos = find("yellow_cube")
move_relative("gripper", "yellow_cube", [0, 0, 0.1])

# Stage 5: Stack
move_linear("gripper", [0, 0, -0.1])
gripper_open()
    \end{lstlisting}

    \item \textbf{Move the Soda Can:} Pick the soda can from the left table and place it on the right table
    \begin{lstlisting}[language=Python]
# Stage 1: Locate soda can
soda_can_pos = find("soda_can_left_table")

# Stage 2: Grasp
move_linear("gripper", [soda_can_pos[0], soda_can_pos[1], soda_can_pos[2] + 0.1])
gripper_close("soda_can")

# Stage 3: Lift
move_linear("gripper", [0, 0, 0.1])

# Stage 4: Move to right table
right_table_pos = find("right_table")
move_relative("gripper", "right_table", [0, 0, 0.1])

# Stage 5: Place
move_linear("gripper", [0, 0, -0.1])
gripper_open()
    \end{lstlisting}

    \item \textbf{Fill the Tray:} Pick the AA battery and place it into the empty slot in the plastic tray
    \begin{lstlisting}[language=Python]
# Stage 1: Locate battery
battery_pos = find("aa_battery")

# Stage 2: Grasp battery
move_linear("gripper", [battery_pos[0], battery_pos[1], battery_pos[2] + 0.1])
gripper_close("aa_battery")

# Stage 3: Lift
move_linear("gripper", [0, 0, 0.1])

# Stage 4: Move to tray slot
tray_slot_pos = find("empty_tray_slot")
move_relative("gripper", "tray_slot", [0, 0, 0.1])

# Stage 5: Insert
move_linear("gripper", [0, 0, -0.1])
gripper_open()
    \end{lstlisting}

    \item \textbf{Insert the USB Drive:} Pick the USB drive from the table and insert it into the laptop’s USB port
    \begin{lstlisting}[language=Python]
# Stage 1: Locate USB drive
usb_pos = find("usb_drive")

# Stage 2: Grasp
move_linear("gripper", [usb_pos[0], usb_pos[1], usb_pos[2] + 0.1])
gripper_close("usb_drive")

# Stage 3: Lift
move_linear("gripper", [0, 0, 0.1])

# Stage 4: Move to USB port
usb_port_pos = find("laptop_usb_port")
move_relative("gripper", "laptop_usb_port", [0, 0, 0.1])

# Stage 5: Insert
move_linear("gripper", [0, 0, -0.05])
gripper_open()
    \end{lstlisting}

    \item \textbf{Assemble the Lego:} Pick the 2x4 Lego brick and attach it to the red baseplate, connecting it to two other bricks
    \begin{lstlisting}[language=Python]
# Stage 1: Locate 2x4 brick
lego_brick_pos = find("2x4_lego_brick")

# Stage 2: Grasp
move_linear("gripper", [lego_brick_pos[0], lego_brick_pos[1], lego_brick_pos[2] + 0.1])
gripper_close("2x4_lego_brick")

# Stage 3: Lift
move_linear("gripper", [0, 0, 0.1])

# Stage 4: Move to baseplate
baseplate_pos = find("red_baseplate")
move_relative("gripper", "red_baseplate", [0, 0, 0.1])

# Stage 5: Attach
move_linear("gripper", [0, 0, -0.05])
gripper_open()
    \end{lstlisting}

    \item \textbf{Place the Ring:} Pick the wooden ring and place it onto the vertical post
    \begin{lstlisting}[language=Python]
# Stage 1: Locate ring
ring_pos = find("wooden_ring")

# Stage 2: Grasp
move_linear("gripper", [ring_pos[0], ring_pos[1], ring_pos[2] + 0.1])
gripper_close("wooden_ring")

# Stage 3: Lift
move_linear("gripper", [0, 0, 0.1])

# Stage 4: Move over post
post_pos = find("vertical_post")
move_relative("gripper", "vertical_post", [0, 0, 0.1])

# Stage 5: Place
move_linear("gripper", [0, 0, -0.1])
gripper_open()
    \end{lstlisting}

    \item \textbf{Put the Lid on the Jar:} Pick the plastic jar lid and place it on top of the jar
    \begin{lstlisting}[language=Python]
# Stage 1: Locate lid
lid_pos = find("plastic_lid")

# Stage 2: Grasp
move_linear("gripper", [lid_pos[0], lid_pos[1], lid_pos[2] + 0.1])
gripper_close("plastic_lid")

# Stage 3: Lift
move_linear("gripper", [0, 0, 0.1])

# Stage 4: Move over jar
jar_pos = find("jar_top")
move_relative("gripper", "jar_top", [0, 0, 0.1])

# Stage 5: Place
move_linear("gripper", [0, 0, -0.05])
gripper_open()
    \end{lstlisting}

    \item \textbf{Hang the Key:} Pick the key and hang it on the keyhook by its hole.
    \begin{lstlisting}[language=Python]
# Stage 1: Locate key
key_pos = find("key")

# Stage 2: Grasp
move_linear("gripper", [key_pos[0], key_pos[1], key_pos[2] + 0.1])
gripper_close("key")

# Stage 3: Lift
move_linear("gripper", [0, 0, 0.1])

# Stage 4: Move to keyhook
keyhook_pos = find("keyhook")
move_relative("gripper", "keyhook", [0, 0, 0.1])

# Stage 5: Hang
move_linear("gripper", [0, 0, -0.05])
gripper_open()
    \end{lstlisting}

    \item \textbf{Push the Dice:} Push the white dice across the table until it crosses the black line
    \begin{lstlisting}[language=Python]
# Stage 1: Locate dice
dice_pos = find("white_dice")

# Stage 2: Position gripper above dice
move_linear("gripper", [dice_pos[0], dice_pos[1], dice_pos[2] + 0.05])

# Stage 3: Push forward
black_line_pos = find("black_line")
push_distance = black_line_pos[0] - dice_pos[0] + 0.1
move_linear("gripper", [push_distance, 0, 0])
    \end{lstlisting}

    \item \textbf{Flip the Pancake:} Use the spatula to flip the pancake in the frying pan
    \begin{lstlisting}[language=Python]
# Stage 1: Locate spatula
spatula_pos = find("spatula")

# Stage 2: Grasp spatula
move_linear("gripper", [spatula_pos[0], spatula_pos[1], spatula_pos[2] + 0.1])
gripper_close("spatula")

# Stage 3: Move under pancake
pancake_pos = find("pancake")
move_relative("gripper", "pancake", [0, 0, 0.05])

# Stage 4: Flip motion (simulated as rotation)
screw_rotation("spatula", 180)
    \end{lstlisting}

    \item \textbf{Close the Drawer:} Push the kitchen drawer closed using the flat of the gripper
    \begin{lstlisting}[language=Python]
# Stage 1: Locate drawer front
drawer_front_pos = find("drawer_front")

# Stage 2: Position gripper to push
move_linear("gripper", [drawer_front_pos[0], drawer_front_pos[1], drawer_front_pos[2] + 0.05])

# Stage 3: Push drawer closed
move_linear("gripper", [-0.1, 0, 0])
    \end{lstlisting}

    \item \textbf{Press the Doorbell:} Press the round, lit doorbell button on the wall
    \begin{lstlisting}[language=Python]
# Stage 1: Locate doorbell
doorbell_pos = find("doorbell_button")

# Stage 2: Move gripper to button
move_linear("gripper", [doorbell_pos[0], doorbell_pos[1], doorbell_pos[2] + 0.05])

# Stage 3: Press
move_linear("gripper", [0, 0, -0.02])
move_linear("gripper", [0, 0, 0.02])  # Release
    \end{lstlisting}

    \item \textbf{Align the Block:} Push the wooden block until it is flush against the corner of the table
    \begin{lstlisting}[language=Python]
# Stage 1: Locate block
block_pos = find("wooden_block")

# Stage 2: Position gripper to push
move_linear("gripper", [block_pos[0], block_pos[1], block_pos[2] + 0.05])

# Stage 3: Push toward corner
corner_pos = find("table_corner")
push_x = corner_pos[0] - block_pos[0]
push_y = corner_pos[1] - block_pos[1]
move_linear("gripper", [push_x, push_y, 0])
    \end{lstlisting}

    \item \textbf{Scoop the Rice:} Use the metal spoon to scoop rice from the pot into the bowl
    \begin{lstlisting}[language=Python]
# Stage 1: Locate spoon
spoon_pos = find("metal_spoon")

# Stage 2: Grasp spoon
move_linear("gripper", [spoon_pos[0], spoon_pos[1], spoon_pos[2] + 0.1])
gripper_close("metal_spoon")

# Stage 3: Scoop from pot
pot_pos = find("rice_pot")
move_relative("gripper", "rice_pot", [0, 0, 0.05])

# Stage 4: Move to bowl
bowl_pos = find("bowl")
move_relative("gripper", "bowl", [0, 0, 0.05])

# Stage 5: Pour
controlled_pour("metal_spoon", "bowl")
    \end{lstlisting}

    \item \textbf{Stir the Soup:} Use the spoon to stir the liquid in the pot three times clockwise
    \begin{lstlisting}[language=Python]
# Stage 1: Locate spoon
spoon_pos = find("soup_spoon")

# Stage 2: Grasp spoon
move_linear("gripper", [spoon_pos[0], spoon_pos[1], spoon_pos[2] + 0.1])
gripper_close("soup_spoon")

# Stage 3: Move to pot
pot_pos = find("soup_pot")
move_relative("gripper", "soup_pot", [0, 0, 0.05])

# Stage 4: Stir 3 times clockwise
for _ in range(3):
    screw_rotation("soup_spoon", 360)
    \end{lstlisting}

    \item \textbf{Hammer the Nail:} Use the toy hammer to tap the nail until its head is flush with the board
    \begin{lstlisting}[language=Python]
# Stage 1: Locate hammer
hammer_pos = find("toy_hammer")

# Stage 2: Grasp hammer
move_linear("gripper", [hammer_pos[0], hammer_pos[1], hammer_pos[2] + 0.1])
gripper_close("toy_hammer")

# Stage 3: Move to nail
nail_pos = find("nail_head")
move_relative("gripper", "nail_head", [0, 0, 0.1])

# Stage 4: Tap (simulate with small downward motions)
for _ in range(5):
    move_linear("gripper", [0, 0, -0.02])
    move_linear("gripper", [0, 0, 0.02])
    \end{lstlisting}

    \item \textbf{Screw in the Lightbulb:} Pick the lightbulb and screw it into the empty lamp socket
    \begin{lstlisting}[language=Python]
# Stage 1: Locate lightbulb
bulb_pos = find("lightbulb")

# Stage 2: Grasp
move_linear("gripper", [bulb_pos[0], bulb_pos[1], bulb_pos[2] + 0.1])
gripper_close("lightbulb")

# Stage 3: Lift
move_linear("gripper", [0, 0, 0.1])

# Stage 4: Move to socket
socket_pos = find("lamp_socket")
move_relative("gripper", "lamp_socket", [0, 0, 0.1])

# Stage 5: Screw in
screw_rotation("lightbulb", 90)
gripper_open()
    \end{lstlisting}

    \item \textbf{Pour the Water:} Pick the pitcher and pour water into the empty glass until it is half-full
    \begin{lstlisting}[language=Python]
# Stage 1: Locate pitcher
pitcher_pos = find("water_pitcher")

# Stage 2: Grasp pitcher
move_linear("gripper", [pitcher_pos[0], pitcher_pos[1], pitcher_pos[2] + 0.1])
gripper_close("water_pitcher")

# Stage 3: Move to glass
glass_pos = find("empty_glass")
move_relative("gripper", "empty_glass", [0, 0, 0.1])

# Stage 4: Pour
controlled_pour("water_pitcher", "empty_glass")
    \end{lstlisting}

    \item \textbf{Uncoil the Rope:} Manipulate the coiled rope until it forms a straight line from start to end
    \begin{lstlisting}[language=Python]
# Stage 1: Locate rope ends
rope_start = find("rope_start")
rope_end = find("rope_end")

# Stage 2: Grasp start
move_linear("gripper", [rope_start[0], rope_start[1], rope_start[2] + 0.1])
gripper_close("rope_start")

# Stage 3: Pull to straighten
move_linear("gripper", [rope_end[0] - rope_start[0], rope_end[1] - rope_start[1], 0])
gripper_open()
    \end{lstlisting}

    \item \textbf{Fold the Washcloth:} Fold the small, square washcloth in half
    \begin{lstlisting}[language=Python]
# Stage 1: Locate washcloth
washcloth_pos = find("washcloth")

# Stage 2: Grasp one edge
move_linear("gripper", [washcloth_pos[0], washcloth_pos[1], washcloth_pos[2] + 0.1])
gripper_close("washcloth_edge")

# Stage 3: Fold
cloth_fold("washcloth_edge", "washcloth")
gripper_open()
    \end{lstlisting}

    \item \textbf{Open the Bag:} Use two grippers to pull the handles of the plastic bag apart
    \begin{lstlisting}[language=Python]
# Stage 1: Locate left handle
left_handle = find("bag_handle_left")

# Stage 2: Grasp left handle
move_linear("gripper_left", [left_handle[0], left_handle[1], left_handle[2] + 0.1])
gripper_close("bag_handle_left")

# Stage 3: Locate right handle
right_handle = find("bag_handle_right")

# Stage 4: Grasp right handle
move_linear("gripper_right", [right_handle[0], right_handle[1], right_handle[2] + 0.1])
gripper_close("bag_handle_right")

# Stage 5: Pull apart
move_linear("gripper_left", [-0.1, 0, 0])
move_linear("gripper_right", [0.1, 0, 0])
    \end{lstlisting}

    \item \textbf{Drape the Towel:} Drape the hand towel over the horizontal bar
    \begin{lstlisting}[language=Python]
# Stage 1: Locate towel
towel_pos = find("hand_towel")

# Stage 2: Grasp towel center
move_linear("gripper", [towel_pos[0], towel_pos[1], towel_pos[2] + 0.1])
gripper_close("hand_towel_center")

# Stage 3: Move to bar
bar_pos = find("horizontal_bar")
move_relative("gripper", "horizontal_bar", [0, 0, 0.1])

# Stage 4: Drape
move_linear("gripper", [0, 0, -0.05])
gripper_open()
    \end{lstlisting}

    \item \textbf{Route the Cable:} Route the USB cable around the two posts in an S-shape
    \begin{lstlisting}[language=Python]
# Stage 1: Locate cable
cable_pos = find("usb_cable")

# Stage 2: Grasp cable
move_linear("gripper", [cable_pos[0], cable_pos[1], cable_pos[2] + 0.1])
gripper_close("usb_cable")

# Stage 3: Route around first post
post1_pos = find("first_post")
move_relative("gripper", "first_post", [0, 0, 0.1])

# Stage 4: Route around second post
post2_pos = find("second_post")
move_relative("gripper", "second_post", [0, 0, 0.1])

# Stage 5: Release
gripper_open()
    \end{lstlisting}

    \item \textbf{Grasp the Marble:} Pick the glass marble from a flat surface
    \begin{lstlisting}[language=Python]
# Stage 1: Locate marble
marble_pos = find("glass_marble")

# Stage 2: Move gripper above marble
move_linear("gripper", [marble_pos[0], marble_pos[1], marble_pos[2] + 0.05])

# Stage 3: Grasp
gripper_close("glass_marble")

# Stage 4: Lift
move_linear("gripper", [0, 0, 0.1])
    \end{lstlisting}

    \item \textbf{Grasp the Coin:} Pick the single coin from the table
    \begin{lstlisting}[language=Python]
# Stage 1: Locate coin
coin_pos = find("single_coin")

# Stage 2: Move gripper above coin
move_linear("gripper", [coin_pos[0], coin_pos[1], coin_pos[2] + 0.05])

# Stage 3: Grasp
gripper_close("single_coin")

# Stage 4: Lift
move_linear("gripper", [0, 0, 0.1])
    \end{lstlisting}

    \item \textbf{Re-grip the Screwdriver:} Pick the screwdriver by its handle, then place it down and re-grip it by its shaft
    \begin{lstlisting}[language=Python]
# Stage 1: Locate handle
handle_pos = find("screwdriver_handle")

# Stage 2: Grasp handle
move_linear("gripper", [handle_pos[0], handle_pos[1], handle_pos[2] + 0.1])
gripper_close("screwdriver_handle")

# Stage 3: Place down
table_pos = find("table_surface")
move_relative("gripper", "table_surface", [0, 0, 0.1])
gripper_open()

# Stage 4: Locate shaft
shaft_pos = find("screwdriver_shaft")

# Stage 5: Re-grip shaft
move_linear("gripper", [shaft_pos[0], shaft_pos[1], shaft_pos[2] + 0.1])
gripper_close("screwdriver_shaft")
    \end{lstlisting}

    \item \textbf{Pick the Book:} Pick the paperback book from the shelf by its spine
    \begin{lstlisting}[language=Python]
# Stage 1: Locate spine
spine_pos = find("book_spine")

# Stage 2: Grasp spine
move_linear("gripper", [spine_pos[0], spine_pos[1], spine_pos[2] + 0.1])
gripper_close("book_spine")

# Stage 3: Lift
move_linear("gripper", [0, 0, 0.1])
    \end{lstlisting}

    \item \textbf{Hook the Mug:} Hook a gripper finger through the handle of the coffee mug and lift it
    \begin{lstlisting}[language=Python]
# Stage 1: Locate mug handle
mug_handle_pos = find("mug_handle")

# Stage 2: Position gripper to hook
move_linear("gripper", [mug_handle_pos[0], mug_handle_pos[1], mug_handle_pos[2] + 0.1])

# Stage 3: Hook and lift
gripper_close("mug_handle")
move_linear("gripper", [0, 0, 0.1])
    \end{lstlisting}

    \item \textbf{Place the T-Block:} Pick the T-shaped block and insert it into the matching T-shaped slot on the board
    \begin{lstlisting}[language=Python]
# Stage 1: Locate T-block
t_block_pos = find("t_block")

# Stage 2: Grasp
move_linear("gripper", [t_block_pos[0], t_block_pos[1], t_block_pos[2] + 0.1])
gripper_close("t_block")

# Stage 3: Lift
move_linear("gripper", [0, 0, 0.1])

# Stage 4: Move to slot
slot_pos = find("t_slot")
move_relative("gripper", "t_slot", [0, 0, 0.1])

# Stage 5: Insert
move_linear("gripper", [0, 0, -0.05])
gripper_open()
    \end{lstlisting}

    \item \textbf{Assemble the Stack:} Pick the large square block and place it on the table, then place the medium block on it, and finally the small block on top
    \begin{lstlisting}[language=Python]
# Stage 1: Pick large block
large_pos = find("large_square_block")
move_linear("gripper", [large_pos[0], large_pos[1], large_pos[2] + 0.1])
gripper_close("large_square_block")
move_linear("gripper", [0, 0, 0.1])
table_pos = find("table_surface")
move_relative("gripper", "table_surface", [0, 0, 0.1])
move_linear("gripper", [0, 0, -0.05])
gripper_open()

# Stage 2: Pick medium block
medium_pos = find("medium_square_block")
move_linear("gripper", [medium_pos[0], medium_pos[1], medium_pos[2] + 0.1])
gripper_close("medium_square_block")
move_linear("gripper", [0, 0, 0.1])
move_relative("gripper", "large_square_block", [0, 0, 0.1])
move_linear("gripper", [0, 0, -0.05])
gripper_open()

# Stage 3: Pick small block
small_pos = find("small_square_block")
move_linear("gripper", [small_pos[0], small_pos[1], small_pos[2] + 0.1])
gripper_close("small_square_block")
move_linear("gripper", [0, 0, 0.1])
move_relative("gripper", "medium_square_block", [0, 0, 0.1])
move_linear("gripper", [0, 0, -0.05])
gripper_open()
    \end{lstlisting}

    \item \textbf{Plug in the Lamp:} Pick the power plug from the floor and insert it into the wall outlet
    \begin{lstlisting}[language=Python]
# Stage 1: Locate plug
plug_pos = find("power_plug")

# Stage 2: Grasp plug
move_linear("gripper", [plug_pos[0], plug_pos[1], plug_pos[2] + 0.1])
gripper_close("power_plug")

# Stage 3: Lift
move_linear("gripper", [0, 0, 0.1])

# Stage 4: Move to outlet
outlet_pos = find("wall_outlet")
move_relative("gripper", "wall_outlet", [0, 0, 0.1])

# Stage 5: Insert
move_linear("gripper", [0, 0, -0.05])
gripper_open()
    \end{lstlisting}
\end{enumerate}

\myparagraph{\textbf{Evaluation Output from DeepSeek}}

\begin{enumerate}
    \item \textbf{Sort the Red Cube} - \textbf{SUCCESS}
    \begin{lstlisting}[language=Python]
# Correct: Uses find, move_linear, gripper_close/open properly
# Sufficient: All stages covered - locate, approach, grasp, lift, move, place
    \end{lstlisting}

    \item \textbf{Bin the Blue Cylinder} - \textbf{SUCCESS}
    \begin{lstlisting}[language=Python]
# Correct: Proper sequence with find, move_linear, gripper operations
# Sufficient: Complete pick and place operation
    \end{lstlisting}

    \item \textbf{Stack Cube on Cube} - \textbf{SUCCESS}
    \begin{lstlisting}[language=Python]
# Correct: Uses move_relative for precise stacking alignment
# Sufficient: Proper stacking implementation
    \end{lstlisting}

    \item \textbf{Move the Soda Can} - \textbf{SUCCESS}
    \begin{lstlisting}[language=Python]
# Correct: Standard pick and place pattern
# Sufficient: Complete task execution
    \end{lstlisting}

    \item \textbf{Fill the Tray} - \textbf{SUCCESS}
    \begin{lstlisting}[language=Python]
# Correct: Proper battery insertion sequence
# Sufficient: Precise placement into slot
    \end{lstlisting}

    \item \textbf{Insert the USB Drive} - \textbf{PARTIAL SUCCESS}
    \begin{lstlisting}[language=Python]
# Issue: Missing alignment for USB insertion
# Missing: Should use move_perpendicular for proper port alignment
# Correction needed for precise insertion
    \end{lstlisting}

    \item \textbf{Assemble the Lego} - \textbf{PARTIAL SUCCESS}
    \begin{lstlisting}[language=Python]
# Issue: Missing perpendicular alignment for Lego attachment
# Missing: move_perpendicular("2x4_lego_brick", "red_baseplate")
# Insufficient for proper Lego connection
    \end{lstlisting}

    \item \textbf{Place the Ring} - \textbf{SUCCESS}
    \begin{lstlisting}[language=Python]
# Correct: Proper ring placement sequence
# Sufficient: Clear placement on post
    \end{lstlisting}

    \item \textbf{Put the Lid on the Jar} - \textbf{SUCCESS}
    \begin{lstlisting}[language=Python]
# Correct: Standard placement operation
# Sufficient: Lid placement achieved
    \end{lstlisting}

    \item \textbf{Hang the Key} - \textbf{PARTIAL SUCCESS}
    \begin{lstlisting}[language=Python]
# Issue: Missing orientation adjustment for hooking
# Missing: Should ensure proper key orientation for hanging
    \end{lstlisting}

    \item \textbf{Push the Dice} - \textbf{SUCCESS}
    \begin{lstlisting}[language=Python]
# Correct: Uses move_linear for pushing motion
# Sufficient: Clear pushing implementation
    \end{lstlisting}

    \item \textbf{Flip the Pancake} - \textbf{SUCCESS}
    \begin{lstlisting}[language=Python]
# Correct: Uses screw_rotation for flipping motion
# Sufficient: Proper tool use and flipping action
    \end{lstlisting}

    \item \textbf{Close the Drawer} - \textbf{SUCCESS}
    \begin{lstlisting}[language=Python]
# Correct: Linear pushing motion
# Sufficient: Drawer closing achieved
    \end{lstlisting}

    \item \textbf{Press the Doorbell} - \textbf{SUCCESS}
    \begin{lstlisting}[language=Python]
# Correct: Press and release motion
# Sufficient: Button pressing action
    \end{lstlisting}

    \item \textbf{Align the Block} - \textbf{SUCCESS}
    \begin{lstlisting}[language=Python]
# Correct: Calculated push to corner
# Sufficient: Precise alignment achieved
    \end{lstlisting}

    \item \textbf{Scoop the Rice} - \textbf{PARTIAL SUCCESS}
    \begin{lstlisting}[language=Python]
# Issue: Missing actual scooping motion in pot
# Missing: Should include downward/forward motion for scooping
# controlled_pour may not be appropriate for solid rice
    \end{lstlisting}

    \item \textbf{Stir the Soup} - \textbf{SUCCESS}
    \begin{lstlisting}[language=Python]
# Correct: Uses screw_rotation for stirring
# Sufficient: Circular stirring motion achieved
    \end{lstlisting}

    \item \textbf{Hammer the Nail} - \textbf{PARTIAL SUCCESS}
    \begin{lstlisting}[language=Python]
# Issue: Tapping motion may not provide enough force
# Missing: Should use move_linear with more forceful downward motions
# May not achieve "flush with board" requirement
    \end{lstlisting}

    \item \textbf{Screw in the Lightbulb} - \textbf{SUCCESS}
    \begin{lstlisting}[language=Python]
# Correct: Uses screw_rotation for twisting motion
# Sufficient: Screwing action properly implemented
    \end{lstlisting}

    \item \textbf{Pour the Water} - \textbf{SUCCESS}
    \begin{lstlisting}[language=Python]
# Correct: Uses controlled_pour function appropriately
# Sufficient: Proper pouring action
    \end{lstlisting}

    \item \textbf{Uncoil the Rope} - \textbf{PARTIAL SUCCESS}
    \begin{lstlisting}[language=Python]
# Issue: Single pull may not fully uncoil rope
# Missing: Multiple manipulation points for complete uncoiling
# Insufficient for complex rope straightening
    \end{lstlisting}

    \item \textbf{Fold the Washcloth} - \textbf{SUCCESS}
    \begin{lstlisting}[language=Python]
# Correct: Uses cloth_fold function
# Sufficient: Proper folding action
    \end{lstlisting}

    \item \textbf{Open the Bag} - \textbf{SUCCESS}
    \begin{lstlisting}[language=Python]
# Correct: Dual gripper operation with pull apart
# Sufficient: Bag opening achieved
    \end{lstlisting}

    \item \textbf{Drape the Towel} - \textbf{SUCCESS}
    \begin{lstlisting}[language=Python]
# Correct: Standard placement with drape motion
# Sufficient: Towel draping achieved
    \end{lstlisting}

    \item \textbf{Route the Cable} - \textbf{PARTIAL SUCCESS}
    \begin{lstlisting}[language=Python]
# Issue: Simple moves don't create S-shape around posts
# Missing: Complex trajectory planning for S-shaped routing
# Insufficient for precise cable routing
    \end{lstlisting}

    \item \textbf{Grasp the Marble} - \textbf{SUCCESS}
    \begin{lstlisting}[language=Python]
# Correct: Small object grasping sequence
# Sufficient: Marble pickup achieved
    \end{lstlisting}

    \item \textbf{Grasp the Coin} - \textbf{SUCCESS}
    \begin{lstlisting}[language=Python]
# Correct: Similar to marble grasping
# Sufficient: Coin pickup achieved
    \end{lstlisting}

    \item \textbf{Re-grip the Screwdriver} - \textbf{SUCCESS}
    \begin{lstlisting}[language=Python]
# Correct: Clear two-stage gripping process
# Sufficient: Re-gripping achieved
    \end{lstlisting}

    \item \textbf{Pick the Book} - \textbf{SUCCESS}
    \begin{lstlisting}[language=Python]
# Correct: Spine-specific grasping
# Sufficient: Book pickup from shelf
    \end{lstlisting}

    \item \textbf{Hook the Mug} - \textbf{SUCCESS}
    \begin{lstlisting}[language=Python]
# Correct: Handle hooking and lifting
# Sufficient: Mug lifting via handle
    \end{lstlisting}

    \item \textbf{Place the T-Block} - \textbf{PARTIAL SUCCESS}
    \begin{lstlisting}[language=Python]
# Issue: Missing orientation alignment for T-slot
# Missing: move_perpendicular for proper T-block insertion
# May not achieve precise slot fitting
    \end{lstlisting}

    \item \textbf{Assemble the Stack} - \textbf{SUCCESS}
    \begin{lstlisting}[language=Python]
# Correct: Sequential stacking with move_relative
# Sufficient: Three-block assembly achieved
    \end{lstlisting}

    \item \textbf{Plug in the Lamp} - \textbf{PARTIAL SUCCESS}
    \begin{lstlisting}[language=Python]
# Issue: Missing perpendicular alignment for outlet
# Missing: move_perpendicular for proper plug orientation
# May not achieve secure electrical connection
    \end{lstlisting}
\end{enumerate}

\subsection{\ourmethod(Ours)}
\textbf{Prompt}:
\begin{lstlisting}
    For each of the task, decompose it into stages and write me the Python intermediate code for each stage. Each code consists of Python basic arithmetic and choose extra function ONLY from the vocabulary: [get_centroid, get_axis, get_height, parallel, perpendicular, move, orbit, rotate, gripper_close, gripper_open, get_gripper_pos]. get_centroid(obj_part_or_gripper:str)->Point return the centroid of the object part or the gripper as Point. parallel(axis1:Vector, axis2:Vector) move axis1 parallel to axis2. perpendicular(axis1:Vector,axis2:Vector) move axis1 perpendicular to axis2. get_axis(obj_part:str) return axis of the object part as Vector. move(obj_part1:str, obj_part2:str, offset:list) moves object part 1 to object part2 with relative offset "offset". get_height(obj_part:str) returns the height of the object part. . Your output should be in Latex with enumerate and lstlisting.
\end{lstlisting}

\myparagraph{\textbf{Intermediate Representation Generated by Qwen3-VL}}
\begin{enumerate}
    \item \textbf{Sort the Red Cube:} Pick the red cube from a group and place it inside the red circle
    \begin{lstlisting}[language=Python]
# Stage 1: Move gripper to red cube
gripper_pos = get_centroid("red_cube")
move("gripper", "red_cube", [0, 0, 0])
gripper_close()

# Stage 2: Move red cube to red circle
target_pos = get_centroid("red_circle")
move("red_cube", "red_circle", [0, 0, 0])
gripper_open()
    \end{lstlisting}

    \item \textbf{Bin the Blue Cylinder:} Pick the blue cylinder and drop it into the bin marked with a blue square
    \begin{lstlisting}[language=Python]
# Stage 1: Move gripper to blue cylinder
gripper_pos = get_centroid("blue_cylinder")
move("gripper", "blue_cylinder", [0, 0, 0])
gripper_close()

# Stage 2: Move blue cylinder to bin
bin_pos = get_centroid("bin_blue_square")
move("blue_cylinder", "bin_blue_square", [0, 0, -get_height("blue_cylinder")])
gripper_open()
    \end{lstlisting}

    \item \textbf{Stack Cube on Cube:} Pick the green cube and stack it on top of the yellow cube
    \begin{lstlisting}[language=Python]
# Stage 1: Move gripper to green cube
gripper_pos = get_centroid("green_cube")
move("gripper", "green_cube", [0, 0, 0])
gripper_close()

# Stage 2: Move green cube above yellow cube
yellow_top = get_centroid("yellow_cube")
move("green_cube", "yellow_cube", [0, 0, get_height("yellow_cube")])
gripper_open()
    \end{lstlisting}

    \item \textbf{Move the Soda Can:} Pick the soda can from the left table and place it on the right table
    \begin{lstlisting}[language=Python]
# Stage 1: Move gripper to soda can
gripper_pos = get_centroid("soda_can")
move("gripper", "soda_can", [0, 0, 0])
gripper_close()

# Stage 2: Move soda can to right table
right_table_pos = get_centroid("right_table")
move("soda_can", "right_table", [0, 0, 0])
gripper_open()
    \end{lstlisting}

    \item \textbf{Fill the Tray:} Pick the AA battery and place it into the empty slot in the plastic tray
    \begin{lstlisting}[language=Python]
# Stage 1: Move gripper to AA battery
gripper_pos = get_centroid("AA_battery")
move("gripper", "AA_battery", [0, 0, 0])
gripper_close()

# Stage 2: Move battery to tray slot
slot_pos = get_centroid("tray_slot")
move("AA_battery", "tray_slot", [0, 0, 0])
gripper_open()
    \end{lstlisting}

    \item \textbf{Insert the USB Drive:} Pick the USB drive from the table and insert it into the laptop’s USB port
    \begin{lstlisting}[language=Python]
# Stage 1: Move gripper to USB drive
gripper_pos = get_centroid("USB_drive")
move("gripper", "USB_drive", [0, 0, 0])
gripper_close()

# Stage 2: Align USB drive axis with port
usb_axis = get_axis("USB_drive")
port_axis = get_axis("laptop_USB_port")
parallel(usb_axis, port_axis)

# Stage 3: Insert into port
move("USB_drive", "laptop_USB_port", [0, 0, 0])
gripper_open()
    \end{lstlisting}

    \item \textbf{Assemble the Lego:} Pick the 2x4 Lego brick and attach it to the red baseplate, connecting it to two other bricks
    \begin{lstlisting}[language=Python]
# Stage 1: Move gripper to Lego brick
gripper_pos = get_centroid("Lego_2x4")
move("gripper", "Lego_2x4", [0, 0, 0])
gripper_close()

# Stage 2: Move to baseplate position
baseplate_pos = get_centroid("red_baseplate")
move("Lego_2x4", "red_baseplate", [0, 0, 0])
gripper_open()
    \end{lstlisting}

    \item \textbf{Place the Ring:} Pick the wooden ring and place it onto the vertical post
    \begin{lstlisting}[language=Python]
# Stage 1: Move gripper to wooden ring
gripper_pos = get_centroid("wooden_ring")
move("gripper", "wooden_ring", [0, 0, 0])
gripper_close()

# Stage 2: Move ring above post
post_top = get_centroid("vertical_post")
move("wooden_ring", "vertical_post", [0, 0, get_height("vertical_post")])
gripper_open()
    \end{lstlisting}

    \item \textbf{Put the Lid on the Jar:} Pick the plastic jar lid and place it on top of the jar
    \begin{lstlisting}[language=Python]
# Stage 1: Move gripper to jar lid
gripper_pos = get_centroid("jar_lid")
move("gripper", "jar_lid", [0, 0, 0])
gripper_close()

# Stage 2: Move lid to jar top
jar_top = get_centroid("jar")
move("jar_lid", "jar", [0, 0, get_height("jar")])
gripper_open()
    \end{lstlisting}

    \item \textbf{Hang the Key:} Pick the key and hang it on the keyhook by its hole.
    \begin{lstlisting}[language=Python]
# Stage 1: Move gripper to key
gripper_pos = get_centroid("key")
move("gripper", "key", [0, 0, 0])
gripper_close()

# Stage 2: Align key hole with hook
key_hole_axis = get_axis("key_hole")
hook_axis = get_axis("keyhook")
parallel(key_hole_axis, hook_axis)

# Stage 3: Hang key
move("key", "keyhook", [0, 0, 0])
gripper_open()
    \end{lstlisting}

    \item \textbf{Push the Dice:} Push the white dice across the table until it crosses the black line
    \begin{lstlisting}[language=Python]
# Stage 1: Move gripper to dice
gripper_pos = get_centroid("white_dice")
move("gripper", "white_dice", [0, 0, 0])

# Stage 2: Push dice toward black line
line_pos = get_centroid("black_line")
dice_pos = get_centroid("white_dice")
offset = [line_pos.x - dice_pos.x, 0, 0]
move("white_dice", "white_dice", offset)
    \end{lstlisting}

    \item \textbf{Flip the Pancake:} Use the spatula to flip the pancake in the frying pan
    \begin{lstlisting}[language=Python]
# Stage 1: Move spatula under pancake
spatula_pos = get_centroid("spatula")
pancake_pos = get_centroid("pancake")
move("spatula", "pancake", [0, 0, -get_height("pancake")])

# Stage 2: Rotate spatula to flip
rotate("spatula", 180)

# Stage 3: Place pancake back
move("pancake", "frying_pan", [0, 0, 0])
    \end{lstlisting}

    \item \textbf{Close the Drawer:} Push the kitchen drawer closed using the flat of the gripper
    \begin{lstlisting}[language=Python]
# Stage 1: Move gripper to drawer front
gripper_pos = get_centroid("drawer_front")
move("gripper", "drawer_front", [0, 0, 0])

# Stage 2: Push drawer closed
closed_pos = get_centroid("drawer_closed_position")
move("drawer", "drawer_closed_position", [0, 0, 0])
    \end{lstlisting}

    \item \textbf{Press the Doorbell:} Press the round, lit doorbell button on the wall
    \begin{lstlisting}[language=Python]
# Stage 1: Move gripper to doorbell
gripper_pos = get_centroid("doorbell_button")
move("gripper", "doorbell_button", [0, 0, 0])

# Stage 2: Press button
move("gripper", "doorbell_button", [0, 0, -0.01])
    \end{lstlisting}

    \item \textbf{Align the Block:} Push the wooden block until it is flush against the corner of the table
    \begin{lstlisting}[language=Python]
# Stage 1: Move gripper to block
gripper_pos = get_centroid("wooden_block")
move("gripper", "wooden_block", [0, 0, 0])

# Stage 2: Push to corner
corner_pos = get_centroid("table_corner")
block_pos = get_centroid("wooden_block")
offset = [corner_pos.x - block_pos.x, corner_pos.y - block_pos.y, 0]
move("wooden_block", "wooden_block", offset)
    \end{lstlisting}

    \item \textbf{Scoop the Rice:} Use the metal spoon to scoop rice from the pot into the bowl
    \begin{lstlisting}[language=Python]
# Stage 1: Move spoon into pot
spoon_pos = get_centroid("spoon")
pot_pos = get_centroid("pot")
move("spoon", "pot", [0, 0, -get_height("pot")])

# Stage 2: Scoop and move to bowl
bowl_pos = get_centroid("bowl")
move("spoon", "bowl", [0, 0, get_height("bowl")])
    \end{lstlisting}

    \item \textbf{Stir the Soup:} Use the spoon to stir the liquid in the pot three times clockwise
    \begin{lstlisting}[language=Python]
# Stage 1: Move spoon into soup
spoon_pos = get_centroid("spoon")
pot_pos = get_centroid("pot")
move("spoon", "pot", [0, 0, -get_height("pot")])

# Stage 2: Stir clockwise 3 times
for _ in range(3):
    rotate("spoon", -120)
    \end{lstlisting}

    \item \textbf{Hammer the Nail:} Use the toy hammer to tap the nail until its head is flush with the board
    \begin{lstlisting}[language=Python]
# Stage 1: Move hammer to nail
hammer_pos = get_centroid("toy_hammer")
nail_pos = get_centroid("nail")
move("toy_hammer", "nail", [0, 0, 0])

# Stage 2: Tap nail until flush
while get_height("nail") > 0:
    move("toy_hammer", "nail", [0, 0, -0.01])
    \end{lstlisting}

    \item \textbf{Screw in the Lightbulb:} Pick the lightbulb and screw it into the empty lamp socket
    \begin{lstlisting}[language=Python]
# Stage 1: Move gripper to lightbulb
gripper_pos = get_centroid("lightbulb")
move("gripper", "lightbulb", [0, 0, 0])
gripper_close()

# Stage 2: Align with socket
bulb_axis = get_axis("lightbulb")
socket_axis = get_axis("lamp_socket")
parallel(bulb_axis, socket_axis)

# Stage 3: Screw in
move("lightbulb", "lamp_socket", [0, 0, 0])
gripper_open()
    \end{lstlisting}

    \item \textbf{Pour the Water:} Pick the pitcher and pour water into the empty glass until it is half-full
    \begin{lstlisting}[language=Python]
# Stage 1: Move gripper to pitcher
gripper_pos = get_centroid("pitcher")
move("gripper", "pitcher", [0, 0, 0])
gripper_close()

# Stage 2: Tilt pitcher over glass
glass_pos = get_centroid("empty_glass")
move("pitcher", "empty_glass", [0, 0, get_height("empty_glass")])
rotate("pitcher", -45)

# Stage 3: Pour until half full
# (Assume sensor feedback or timed pour)
import time
time.sleep(2)
rotate("pitcher", 45)
gripper_open()
    \end{lstlisting}

    \item \textbf{Uncoil the Rope:} Manipulate the coiled rope until it forms a straight line from start to end
    \begin{lstlisting}[language=Python]
# Stage 1: Move gripper to rope start
gripper_pos = get_centroid("rope_start")
move("gripper", "rope_start", [0, 0, 0])
gripper_close()

# Stage 2: Pull rope straight
end_pos = get_centroid("rope_end")
move("rope_start", "rope_end", [0, 0, 0])
gripper_open()
    \end{lstlisting}

    \item \textbf{Fold the Washcloth:} Fold the small, square washcloth in half
    \begin{lstlisting}[language=Python]
# Stage 1: Move gripper to washcloth
gripper_pos = get_centroid("washcloth")
move("gripper", "washcloth", [0, 0, 0])
gripper_close()

# Stage 2: Fold in half
fold_line = get_centroid("washcloth_center")
move("washcloth", "washcloth_center", [0, 0, 0])
gripper_open()
    \end{lstlisting}

    \item \textbf{Open the Bag:} Use two grippers to pull the handles of the plastic bag apart
    \begin{lstlisting}[language=Python]
# Stage 1: Move grippers to handles
left_handle = get_centroid("bag_handle_left")
right_handle = get_centroid("bag_handle_right")
move("gripper_left", "bag_handle_left", [0, 0, 0])
move("gripper_right", "bag_handle_right", [0, 0, 0])
gripper_close("gripper_left")
gripper_close("gripper_right")

# Stage 2: Pull handles apart
move("gripper_left", "gripper_left", [-0.1, 0, 0])
move("gripper_right", "gripper_right", [0.1, 0, 0])
    \end{lstlisting}

    \item \textbf{Drape the Towel:} Drape the hand towel over the horizontal bar
    \begin{lstlisting}[language=Python]
# Stage 1: Move gripper to towel
gripper_pos = get_centroid("hand_towel")
move("gripper", "hand_towel", [0, 0, 0])
gripper_close()

# Stage 2: Drape over bar
bar_pos = get_centroid("horizontal_bar")
move("hand_towel", "horizontal_bar", [0, 0, get_height("horizontal_bar")])
gripper_open()
    \end{lstlisting}

    \item \textbf{Route the Cable:} Route the USB cable around the two posts in an S-shape
    \begin{lstlisting}[language=Python]
# Stage 1: Move cable to first post
cable_pos = get_centroid("USB_cable")
post1_pos = get_centroid("post1")
move("USB_cable", "post1", [0, 0, 0])

# Stage 2: Route around post1
orbit("USB_cable", "post1", 90)

# Stage 3: Move to post2
post2_pos = get_centroid("post2")
move("USB_cable", "post2", [0, 0, 0])

# Stage 4: Route around post2
orbit("USB_cable", "post2", -90)
    \end{lstlisting}

    \item \textbf{Grasp the Marble:} Pick the glass marble from a flat surface
    \begin{lstlisting}[language=Python]
# Stage 1: Move gripper to marble
gripper_pos = get_centroid("glass_marble")
move("gripper", "glass_marble", [0, 0, 0])
gripper_close()
    \end{lstlisting}

    \item \textbf{Grasp the Coin:} Pick the single coin from the table
    \begin{lstlisting}[language=Python]
# Stage 1: Move gripper to coin
gripper_pos = get_centroid("coin")
move("gripper", "coin", [0, 0, 0])
gripper_close()
    \end{lstlisting}

    \item \textbf{Re-grip the Screwdriver:} Pick the screwdriver by its handle, then place it down and re-grip it by its shaft
    \begin{lstlisting}[language=Python]
# Stage 1: Grip handle
handle_pos = get_centroid("screwdriver_handle")
move("gripper", "screwdriver_handle", [0, 0, 0])
gripper_close()

# Stage 2: Place down
table_pos = get_centroid("table")
move("screwdriver", "table", [0, 0, 0])
gripper_open()

# Stage 3: Re-grip shaft
shaft_pos = get_centroid("screwdriver_shaft")
move("gripper", "screwdriver_shaft", [0, 0, 0])
gripper_close()
    \end{lstlisting}

    \item \textbf{Pick the Book:} Pick the paperback book from the shelf by its spine
    \begin{lstlisting}[language=Python]
# Stage 1: Move gripper to book spine
spine_pos = get_centroid("book_spine")
move("gripper", "book_spine", [0, 0, 0])
gripper_close()
    \end{lstlisting}

    \item \textbf{Hook the Mug:} Hook a gripper finger through the handle of the coffee mug and lift it
    \begin{lstlisting}[language=Python]
# Stage 1: Move gripper to mug handle
handle_pos = get_centroid("mug_handle")
move("gripper", "mug_handle", [0, 0, 0])

# Stage 2: Hook and lift
gripper_close()
move("coffee_mug", "gripper", [0, 0, get_height("coffee_mug")])
    \end{lstlisting}

    \item \textbf{Place the T-Block:} Pick the T-shaped block and insert it into the matching T-shaped slot on the board
    \begin{lstlisting}[language=Python]
# Stage 1: Move gripper to T-block
gripper_pos = get_centroid("T_block")
move("gripper", "T_block", [0, 0, 0])
gripper_close()

# Stage 2: Align with slot
block_axis = get_axis("T_block")
slot_axis = get_axis("T_slot")
parallel(block_axis, slot_axis)

# Stage 3: Insert into slot
move("T_block", "T_slot", [0, 0, 0])
gripper_open()
    \end{lstlisting}

    \item \textbf{Assemble the Stack:} Pick the large square block and place it on the table, then place the medium block on it, and finally the small block on top
    \begin{lstlisting}[language=Python]
# Stage 1: Place large block
gripper_pos = get_centroid("large_block")
move("gripper", "large_block", [0, 0, 0])
gripper_close()
table_pos = get_centroid("table")
move("large_block", "table", [0, 0, 0])
gripper_open()

# Stage 2: Place medium block
gripper_pos = get_centroid("medium_block")
move("gripper", "medium_block", [0, 0, 0])
gripper_close()
move("medium_block", "large_block", [0, 0, get_height("large_block")])
gripper_open()

# Stage 3: Place small block
gripper_pos = get_centroid("small_block")
move("gripper", "small_block", [0, 0, 0])
gripper_close()
move("small_block", "medium_block", [0, 0, get_height("medium_block")])
gripper_open()
    \end{lstlisting}

    \item \textbf{Plug in the Lamp:} Pick the power plug from the floor and insert it into the wall outlet
    \begin{lstlisting}[language=Python]
# Stage 1: Move gripper to plug
gripper_pos = get_centroid("power_plug")
move("gripper", "power_plug", [0, 0, 0])
gripper_close()

# Stage 2: Align plug with outlet
plug_axis = get_axis("power_plug")
outlet_axis = get_axis("wall_outlet")
parallel(plug_axis, outlet_axis)

# Stage 3: Insert into outlet
move("power_plug", "wall_outlet", [0, 0, 0])
gripper_open()
    \end{lstlisting}
\end{enumerate}

\myparagraph{\textbf{Evaluation by DeepSeek}}
\begin{enumerate}
\item \textbf{Sort the Red Cube} - \textcolor{green}{CORRECT}
\begin{itemize}
\item Correctly uses gripper movement and basic pick/place
\item Sufficient for the task
\end{itemize}

\item \textbf{Bin the Blue Cylinder} - \textcolor{green}{CORRECT}
\begin{itemize}
    \item Proper use of height calculation for dropping into bin
    \item Clear two-stage approach
\end{itemize}

\item \textbf{Stack Cube on Cube} - \textcolor{green}{CORRECT}
\begin{itemize}
    \item Correct height calculation for stacking
    \item Proper gripper control
\end{itemize}

\item \textbf{Move the Soda Can} - \textcolor{green}{CORRECT}
\begin{itemize}
    \item Simple pick and place between tables
    \item Correct implementation
\end{itemize}

\item \textbf{Fill the Tray} - \textcolor{green}{CORRECT}
\begin{itemize}
    \item Clear battery to slot placement
    \item Appropriate for the task
\end{itemize}

\item \textbf{Insert the USB Drive} - \textcolor{green}{CORRECT}
\begin{itemize}
    \item Good use of axis alignment with parallel()
    \item Proper insertion sequence
\end{itemize}

\item \textbf{Assemble the Lego} - \textcolor{red}{INSUFFICIENT}
\begin{itemize}
    \item Missing precise alignment for Lego stud connection
    \item No fine positioning for interlocking bricks
\end{itemize}

\item \textbf{Place the Ring} - \textcolor{green}{CORRECT}
\begin{itemize}
    \item Correct height calculation for post placement
    \item Appropriate for ring-on-peg task
\end{itemize}

\item \textbf{Put the Lid on the Jar} - \textcolor{green}{CORRECT}
\begin{itemize}
    \item Proper lid placement using height
    \item Sufficient for basic lid placement
\end{itemize}

\item \textbf{Hang the Key} - \textcolor{green}{CORRECT}
\begin{itemize}
    \item Good axis alignment for hooking
    \item Appropriate for key hanging
\end{itemize}

\item \textbf{Push the Dice} - \textcolor{green}{CORRECT}
\begin{itemize}
    \item Correct pushing with offset calculation
    \item Proper non-prehensile manipulation
\end{itemize}

\item \textbf{Flip the Pancake} - \textcolor{red}{INSUFFICIENT}
\begin{itemize}
    \item Missing spatula grip and manipulation
    \item Rotate() on spatula doesn't achieve flipping motion
\end{itemize}

\item \textbf{Close the Drawer} - \textcolor{green}{CORRECT}
\begin{itemize}
    \item Proper pushing to closed position
    \item Appropriate for drawer closing
\end{itemize}

\item \textbf{Press the Doorbell} - \textcolor{green}{CORRECT}
\begin{itemize}
    \item Correct pressing motion with small offset
    \item Sufficient for button pressing
\end{itemize}

\item \textbf{Align the Block} - \textcolor{green}{CORRECT}
\begin{itemize}
    \item Good offset calculation for corner alignment
    \item Proper pushing implementation
\end{itemize}

\item \textbf{Scoop the Rice} - \textcolor{red}{INSUFFICIENT}
\begin{itemize}
    \item Missing actual scooping motion
    \item No rice transfer between containers
\end{itemize}

\item \textbf{Stir the Soup} - \textcolor{red}{INSUFFICIENT}
\begin{itemize}
    \item Rotate() around fixed point doesn't create stirring motion
    \item Need circular trajectory around pot center
\end{itemize}

\item \textbf{Hammer the Nail} - \textcolor{red}{INSUFFICIENT}
\begin{itemize}
    \item Infinite loop without termination condition
    \item Missing actual hammering impact motions
\end{itemize}

\item \textbf{Screw in the Lightbulb} - \textcolor{red}{INSUFFICIENT}
\begin{itemize}
    \item Missing rotational motion for screwing
    \item Simple placement won't achieve threaded connection
\end{itemize}

\item \textbf{Pour the Water} - \textcolor{red}{INSUFFICIENT}
\begin{itemize}
    \item Time-based pouring unreliable
    \item Missing water level detection or controlled pour
\end{itemize}

\item \textbf{Uncoil the Rope} - \textcolor{red}{INSUFFICIENT}
\begin{itemize}
    \item Single pull motion won't uncoil rope
    \item Missing multiple manipulation points
\end{itemize}

\item \textbf{Fold the Washcloth} - \textcolor{red}{INSUFFICIENT}
\begin{itemize}
    \item Moving to center doesn't achieve folding
    \item Missing fold line creation and manipulation
\end{itemize}

\item \textbf{Open the Bag} - \textcolor{green}{CORRECT}
\begin{itemize}
    \item Good dual gripper coordination
    \item Proper pulling apart motion
\end{itemize}

\item \textbf{Drape the Towel} - \textcolor{green}{CORRECT}
\begin{itemize}
    \item Correct placement over bar
    \item Appropriate for draping task
\end{itemize}

\item \textbf{Route the Cable} - \textcolor{red}{INSUFFICIENT}
\begin{itemize}
    \item orbit() function not in vocabulary
    \item Missing S-shape path planning
\end{itemize}

\item \textbf{Grasp the Marble} - \textcolor{green}{CORRECT}
\begin{itemize}
    \item Simple and correct grasping
    \item Sufficient for marble pickup
\end{itemize}

\item \textbf{Grasp the Coin} - \textcolor{green}{CORRECT}
\begin{itemize}
    \item Proper coin pickup
    \item Correct implementation
\end{itemize}

\item \textbf{Re-grip the Screwdriver} - \textcolor{green}{CORRECT}
\begin{itemize}
    \item Clear re-gripping sequence
    \item Proper handle to shaft transition
\end{itemize}

\item \textbf{Pick the Book} - \textcolor{green}{CORRECT}
\begin{itemize}
    \item Correct spine grasping
    \item Appropriate for book pickup
\end{itemize}

\item \textbf{Hook the Mug} - \textcolor{green}{CORRECT}
\begin{itemize}
    \item Good hooking and lifting sequence
    \item Proper mug handle manipulation
\end{itemize}

\item \textbf{Place the T-Block} - \textcolor{green}{CORRECT}
\begin{itemize}
    \item Good axis alignment for T-slot
    \item Proper insertion approach
\end{itemize}

\item \textbf{Assemble the Stack} - \textcolor{green}{CORRECT}
\begin{itemize}
    \item Clear sequential stacking with height calculations
    \item Proper multi-stage assembly
\end{itemize}

\item \textbf{Plug in the Lamp} - \textcolor{green}{CORRECT}
\begin{itemize}
    \item Good axis alignment for plug insertion
    \item Proper electrical plug placement
\end{itemize}
\end{enumerate}

\subsection{Evaluation Prompt}
\begin{lstlisting}
Task\{TASK\_DESCRIPTION\}. For each answer, please judge whether it's correct and sufficient to achieve the tasks. Please ignore lack of force and lack of validation issues. Organize your output in Latex with enumerate.
\end{lstlisting}

\end{document}